\documentclass[sigconf, nonacm]{acmart}
\AtBeginDocument{%
  }
\usepackage{siunitx}
\begin{document}
\title{BiTro: Bidirectional Transfer Learning Enhances Bulk and Spatial Transcriptomics Prediction in Cancer Pathological Images}

\author{Jingkun Yu}
\authornote{Equal contributions.}
\email{2321466165@my.swjtu.edu.cn}
\orcid{1234-5678-9012}
\author{Guangkai Shang}
\authornotemark[1]
\email{sgk200506@my.swjtu.edu.cn}
\affiliation{%
  \institution{SWJTU-Leeds Joint School, Southwest Jiaotong University}
  \city{Chengdu}
  \country{China}
}

\author{Changtao Li}
\email{changtao_li@stu.scu.edu.cn}
\affiliation{%
  \institution{Yu-Yue Pathology Scientific Research Center, Jinfeng Laboratory}
  \city{Chongqing}
  \country{China}
}
\affiliation{%
  \institution{Department of Epidemiology and Medical Statistics, West China School of Public Health and West China Fourth Hospital, Sichuan University}
  \city{Chengdu}
  \country{China}
}

\author{Xun Gong}
\email{gongxun@foxmail.com}
\affiliation{%
  \institution{School of Computing and Artificial Intelligence, Southwest Jiaotong University}
  \city{Chengdu}
  \country{China}
}

\author{Tianrui Li}
\email{trli@swjtu.edu.cn}
\affiliation{%
  \institution{School of Computing and Artificial Intelligence, Southwest Jiaotong University}
  \city{Chengdu}
  \country{China}
}

\author{Yazhou He}
\email{y.he@imperial.ac.uk}
\authornote{Corresponding authors.}
\affiliation{%
  \institution{Yu-Yue Pathology Scientific Research Center, Jinfeng Laboratory}
  \city{Chongqing}
  \country{China}
}
\affiliation{%
  \institution{Department of Epidemiology and Biostatistics, School of Public Health, Imperial College London}
  \city{London}
  \country{UK}
}

\author{Zhipeng Luo}
\authornotemark[2]
\email{zpluo@swjtu.edu.cn}
\affiliation{%
  \institution{School of Computing and Artificial Intelligence, Southwest Jiaotong University}
  \city{Chengdu}
  \country{China}
}

% 必须在 \begin{document} 之后，\maketitle 之前定义这部分
\renewcommand{\shortauthors}{Yu and Shang, et al.}


\begin{abstract}
% Cancer imposes a severe global burden on public health, and integrating pathological imaging with genomic data is crucial for understanding tumor heterogeneity. 
% Models predicting gene expression from Whole Slide Imaging (WSI) either only yield coarse-grained bulk-level results or are prone to overfitting due to insufficient ST data and low sequencing depth. 
% To date, no model has achieved gene expression inference based on single-cell feature-level modeling.
% This study pioneers a new direction of bidirectional transfer between bulk sequencing and ST sequencing, leveraging the distinct advantages of the two data types. It provides insights for future research utilizing cell-level gene expression data.

Cancer pathological analysis requires modeling tumor heterogeneity across multiple modalities, primarily through transcriptomics and whole slide imaging (WSI), along with their spatial relations. 
On one hand, bulk transcriptomics and WSI images are largely available but lack spatial mapping; on the other hand, spatial transcriptomics (ST) data can offer high spatial resolution, yet facing challenges of high cost, low sequencing depth, and limited sample sizes.
Therefore, the data foundation of either side is flawed and has its limit in accurately finding the mapping between the two modalities.
To this end, we propose BiTro, a bidirectional transfer learning framework that can enhance bulk and spatial transcriptomics prediction from pathological images.
Our contributions are twofold. First, we design a universal and transferable model architecture that works for both bulk+WSI and ST data. A major highlight is that we model WSI images on the cellular level to better capture cells' visual features,  morphological phenotypes, and their spatial relations; to map cells' features to their transcriptomics measured in bulk or ST, we adopt multiple instance learning.
Second, by using LoRA, our model can be efficiently transferred between bulk and ST data to exploit their complementary information.
To test our framework, we conducted comprehensive experiments on five cancer datasets. Results demonstrate that 1) our base model can achieve better or competitive performance compared to existing models on bulk or spatial transcriptomics prediction, and 2) transfer learning can further improve the base model's performance. \footnote{Our code is available at \href{https://github.com/yujingkun1/BiTro}{https://github.com/yujingkun1/BiTro}.}

\end{abstract}



%% page.这里就是填写我们自己的引用格式


\maketitle

\section{Introduction}
Cancer remains a major public health challenge and continues to be the leading cause of human deaths globally~\cite{siegel2025cancer}. Cancer pathology is the medical examination of tissue, cells, or body fluids to definitively diagnose cancer and analyze its causes and progression. Thus, in nature, cancer pathological analysis is a comprehensive task and requires the data foundation to be multimodal, fine-grained, and large-scale.
Thanks to the advances in biomedical technologies such as high-throughput gene sequencing and microscopic imaging, the data foundation is being built with tumor transcriptomes and whole slide images (WSI).
A tumor's transcriptome can provide summed gene expressions with wide coverage and high depth. 
Whole slide imaging of high-resolution pathological sections can reveal the morphological characteristics of a tumor's tissues and cells \cite{zarella2019practical,mukhopadhyay2018whole,martinez2020compendium}.
Currently, large-scale paired bulk transcriptomics and WSI images of various cancer cohorts are publicly available from The Cancer Genome Atlas (TCGA) database \cite{weinstein2013cancer}. Nevertheless, such data modalities are sometimes considered coarse-grained, as they cannot map omics data directly onto the physical structure of tissues, preventing the discovering in tumor's microenvironment.

Recently, spatial transcriptomics (ST) has emerged and is mitigating the data gap. ST combines gene expression sequencing and microscopic imaging technologies and pushes the granularity to the \textit{spot} level. A spot typically consists of dozens of cells and can be both sequenced and imaged. Therefore, ST is a more powerful tool that enables the analysis of cell states, spatial characteristics, and their interactions within the microenvironment. 
However, the ST technology is relatively new and has limitations of high cost and low sequencing depth. For example, the recommended sequencing amount per spot is 25K reads, which is significantly lower than that of bulks, 60M-150M paired-end reads \cite{nagasawa2024spatial}.
For this reason, ST data often have very limited sample sizes and relatively low sequencing accuracy. When used for training prediction models, this can lead to overfitting and poor generalization.

Our ultimate goal is to build accurate mappings from pathological images to gene expression, as demanded by precision cancer pathology. Prior work seems to model bulk-level data (i.e., bulk transcriptomics with paired WSIs) or ST data separately, as this is naturally isolated by different granularities of data. However, as mentioned above, either data foundation has its pros and cons: bulk-level data are rich but are of low resolution and lack spatial mapping; ST data are finer-grained but are of very limited samples. Therefore, this motivates us to use both levels of data to exploit their complementary information when building the mappings. 

To this end, we propose and develop \textbf{BiTro}, a \textbf{Bi}directional \textbf{Tr}ansfer learning framework to enhance \textbf{Tr}anscript\textbf{o}mics prediction in cancer pathological images.
In BiTro's design, it has a base model architecture that can universally work for bulk or spot-level transcriptomics prediction. 
A major contribution is that we model WSI images on the cellular level to extract fine-grained visual and spatial features. Specifically, we first segment cells by using HoverNet~\cite{graham2019hover} or CellViT~\cite{horst2024cellvit} to obtain their raw pixels and spatial coordinates. Then, by using DinoV3~\cite{simeoni2025dinov3}, we extract cells' semantic visual features, which are also used to identify their morphological phenotypes. In parallel, based on cells' coordinates, we can model their microenvironment (local) features and tumor-level (global) features. Lastly, we combine cells' different features and map them to their bulk or spot transcriptomics by multiple instance learning.
Another major contribution is that we enhance our base model's performance through bidirectional transfer learning. That is, if the target is to predict bulk transcriptomics, we pretrain a model with relevant ST data and finetune it with the target bulk+WSI data, and vice versa. The finetuning process is accelerated by LoRA, an efficient finetuning method.
Overall, BiTro's workflow is illustrated in Fig.~\ref{fig:framework_overview}, and the main contributions can be summarized as follows.

\textbullet\ \ We propose and design a universal and transferable model architecture that works for both bulk or spot-level transcriptomics prediction from pathological images. We model WSI images on the cellular level to better capture cells' visual features, morphological phenotypes, and their spatial relations. Results demonstrate that such a base model, even without transfer learning, can achieve better or competitive performance compared to existing models.

\textbullet\ \ We develop BiTro into a bidirectional transfer learning framework, which further improves the original bulk or spot-level transcriptomics prediction. The finetuning process is implemented efficient with LoRA.

\textbullet\ \ We tested our framework by comprehensive experiments on five cancer datasets against seven to nine baselines. Results are very supportive of the aforementioned functions of BiTro and thus have validated its efficacy and robustness.

\section{Related Work}

\subsection{Modeling WSI and Bulk Transcriptomics}

Paired bulk transcriptomics and WSI pathological images of various cancer cohorts are largely available, such as from TCGA \cite{weinstein2013cancer}. This has produced a plethora of work that quantitatively predicts bulk-level gene expressions from WSIs. An early yet representative study was HE2RNA~\cite{schmauch2020deep}, which trained a neural network that can predict the expression levels of over 20k genes based on tiled WSI images. 
Transformer-based models, such as tRNAsformer~\cite{alsaafin2023learning}, exploit the self-attention mechanism, which can effectively model the interactions among slide tiles, thus better capturing the correlation of gene expressions.
More recently, SEQUOIA~\cite{pizurica2024digital} adopted the UNI foundation model \cite{chen2024towards} customized for pathological images to extract visual features. So SEQUOIA can more accurately capture the intrinsic correlation between tissue morphology and molecular features, thus enhancing the prediction power.

\subsection{Modeling based on Spatial Transcriptomics}

The rise of spatial transcriptomics technology is pushing the data granularity to the spot level or subcellular level. Relying on spot-level gene expression data provided by the Visium platform, ST-Net~\cite{he2020integrating} took the lead by adopting a residual network to realize gene expression prediction at the patch level of WSIs. 
Later, HisToGene~\cite{pang2021leveraging} introduced a Vision Transformer (ViT) network to further strengthen the spatial feature learning; subsequent models such as Hist2ST \cite{zeng2022spatial}, THItoGene \cite{jia2023thitogene}, M2OST~\cite{wang2025m2ost} continued to improve the prediction performance by modeling the spatial connections more delicately.
Lately, ultra-high-resolution data at the subcellular level are being generated by the Xenium technology \cite{marco2025optimizing}. Based on this, iStar~\cite{zhang2024inferring} and GHIST~\cite{fu2025spatial} have achieved cell-level gene expression prediction. But, we want to stress that subcellular ST data are currently very few and are of low sequencing depth. Thus, our work focuses on bulk+WSI and spot-level ST data that have more solid data foundations.

\begin{figure*}[t]
  \centering
  \includegraphics[width=\textwidth]{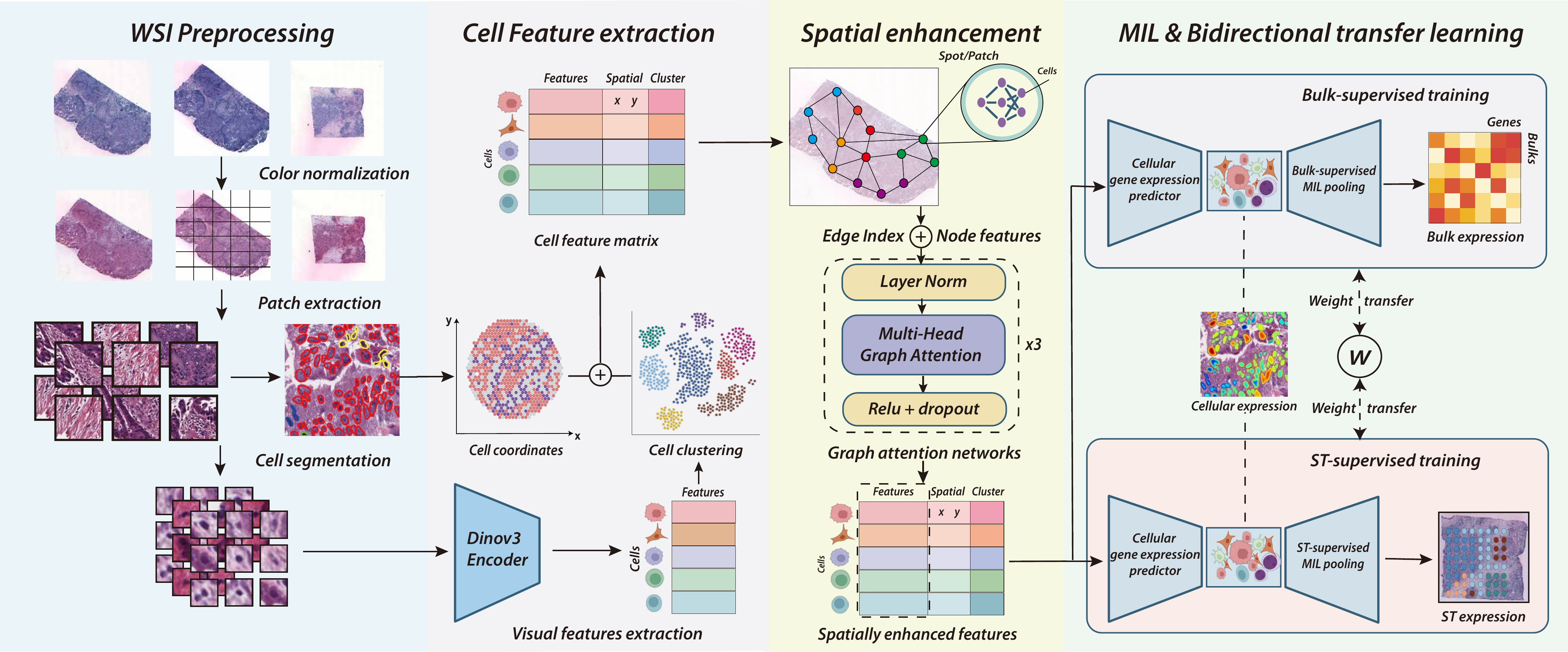}
  \caption{Overview of the BiTro framework. There are four sequential phases: 
  (a) \textbf{WSI preprocessing}: segment a WSI into cells by HoverNet or CellViT; 
  (b) \textbf{Cell feature extraction}: extract cells' visual features via Dinov3, capture cell phenotypes by K-means clustering, and add spatial coordinates; 
  (c) \textbf{Spatial enhancement}: enhance cell features with both local and global spatial relations; 
  (d) \textbf{MIL and Bidirectional transfer learning}: map cellular features to their bulk or spot-level transcriptomics by multiple instance learning. The learning process is enhanced by both levels of data via bidirectional transfer learning.}
  \label{fig:framework_overview}
\end{figure*}

\subsection{Multiple instance learning}

% Currently, cellular spatial transcriptomics data for human cancers are rare, so directly modeling them lacks a solid data foundation. Hence, to reach the cell level, a common practice is to employ multiple instance learning (MIL)\footnote{In our text, we use MIL to broadly represent a class of algorithms that share the same spirit: multiple instance learning, bulk RNA-seq deconvolution, learning from label proportions (classification), and learning from aggregated responses (regression).}  \cite{carbonneau2018multiple}. 

In our model architecture, we work with cell-level features, but the supervision (transcriptomics) is given at the bulk or spot-level. To align features to higher-level supervision, a common practice is to use multiple instance learning (MIL).
MIL is a weakly supervised learning scenario where instance labels are unknown, and labels are only provided to bags, which are sets of instances. In our context, an instance is a cellular feature vector, and a bag is a bulk or a spot of cells whose label is the summed transcriptomics. MIL can be used to learn instance-level or bag-level predictors, and we fall into the second category.
% To ensure the prediction accuracy, MIL requires that the training bags should be sufficiently many, small-sized, and homogeneous \cite{yu2014learning,brahmbhatt2023pac,chen2024general}. 
% Such a principle motivates us to use (bulk RNA-seq, WSI) paired data and ST data together.
MIL has a long history, and early work can be dated back to the 1980s~\cite{lindsay1980applications}. To date, MIL's learning theory has been rigorously studied, such as by \cite{yu2014learning,brahmbhatt2023pac,chen2024general}. The principles motivate us to use bulk+WSI data and ST data together.
Now, MIL has been widely applied to pathological analysis. For example, CoD-MIL~\cite{shi2024cod} uses MIL in the pathologists' diagnostic processes, and \cite{lu2023visual,jaume2024multistain} use MIL in WSI image analysis. Besides, \cite{shaomultiple} conducted a comprehensive study supporting that MIL can be used with transfer learning in pathology.

\section{Methodology}

Our BiTro framework aims to predict gene expression from histology whole slide images (WSIs). The data foundation is a combination of (bulk transcriptomics, WSI) pairs and spot-level spatial transcriptomics (ST).
The whole process is illustrated in Fig.~\ref{fig:framework_overview}. There are four phases: the first three, WSI preprocessing, cell feature extraction, and graph enhancement, serve for cell-level feature extraction; and the last phase is to use multiple instance learning (MIL) to map cellular features to their gene expression, which is further enhanced by transfer learning. The rest of this section details BiTro’s design and implementation.

% Initially, we employ Hover-Net to segment individual nuclei within the WSIs~\cite{graham2019hover}. To capture robust semantic representations, cell-level features are extracted using the DINOv3 foundation model and categorized into distinct phenotypes via K-means clustering~\cite{simeoni2025dinov3} ~\cite{ahmed2020k}. To explicitly model tissue microenvironments, we construct cell graphs where nodes represent cells and edges denote spatial proximity. Graph Attention Networks (GAT) are then utilized to aggregate neighborhood information~\cite{velivckovic2017graph}, thereby enriching cell embeddings with local spatial context. These spatially-enhanced features are processed by a Transformer for Multiple Instance Learning (MIL) to generate slide-level representations. Notably, BiTro adopts a symmetric dual-stream architecture to process WSI-Bulk and ST data in parallel. Finally, a cross-modal transfer learning strategy is introduced to bridge the domain gap, leveraging shared knowledge to boost predictive performance in both modalities. The rest of this section details BiTro’s design and implementation.

\subsection{Preliminaries}

We start with extracting cells' visual features and spatial coordinates from WSI images.
First, each WSI is gridded into patches, representing tissue-level semantics. If a WSI is already spatially resolved, i.e., gene expressions have been measured at spots, we create one patch at each spot; if a WSI is paired only with bulk transcriptomics, the image is divided into spot-sized patches as shown in Fig.~\ref{fig:framework_overview}.
Then, we identify cell centroids within each patch and segment out $N$ tiles $ \{\mathbf{p}_i\}_{i=1}^{N}$ of size $48 \times 48$, where each $\mathbf{p}_i$ represents the $i$-th cell's image pixels. Then, each cell is also associated with a 2D spatial coordinate $\mathbf{c}_i = (x_i, y_i) \in \mathbb{R}^2$, representing its centroid location on the image slide.
Thus, a WSI image patch can be structured as:
\begin{equation}
    \mathbf{X} = \{ (\mathbf{p}_i, \mathbf{c}_i) \}_{i=1}^{N}
\end{equation}
Denote by $G$ the number of genes, and let $\mathbf{y}_{\mathrm{bulk}} $, $ \mathbf{y}_{\mathrm{spot}} \in \mathbb{R}^{G}$ be the gene expression ground truth of a bulk or a spot. Then, a bulk-level dataset can be represented as $\mathcal{D}_{\mathrm{bulk}}=\{\{\mathbf{X}\},\mathbf{y}_{\mathrm{bulk}}\}_{\times{N_\mathrm{b}}}$ and a ST dataset as $\mathcal{D}_{\mathrm{ST}}=\{\mathbf{X},\mathbf{y}_{\mathrm{spot}}\}_{\times{N_\mathrm{s}}}$. The base model is to predict bulk or spot gene expression profiles:
\begin{equation}
    \{\hat{\mathbf{y}}_{\mathrm{bulk}}\}_{\times{N_\mathrm{b}}} = {f}_\theta(\mathcal{D}_{\mathrm{bulk}}),\ \text{or}\ \{\hat{\mathbf{y}}_{\mathrm{spot}}\}_{\times{N_\mathrm{s}}} = {f}_\theta(\mathcal{D}_{\mathrm{ST}})
\end{equation}
If enhanced with transfer learning, the model parameters $\theta$ can be pretrained. For example, if the target task is to predict $\hat{\mathbf{y}}_{\mathrm{bulk}}$, then $\theta$ can be pretrained with a relevant $\mathcal{D}_{\mathrm{ST}}$ dataset and then finetuned with the target $\mathcal{D}_{\mathrm{bulk}}$ data.

\subsection{Cell Feature Extraction}

\paragraph{\textbf{Data preprocessing.}}
WSI images and gene expression data need to be preprocessed, and this includes stain normalization, cell segmentation, gene selection, and gene expression normalization. Details can be found in Appendix~\ref{appendix:preprocess}.

\paragraph{\textbf{Visual feature extraction.}}
We adopt a pretrained Dinov3 (ViT-Large) encoder to extract visual features. 
Each cell's $\mathbf{p}_i$ is originally $48 \times 48$ pixels, but is resized to $224 \times 224$ to accommodate Dino's input requirement. We compute the global average pooling of the last hidden states in Dino and obtain a feature vector $\mathbf{f}_i \in \mathbb{R}^{1024}$.
To reduce redundant information, we apply Principal Component Analysis (PCA) to project $\mathbf{f}_i$ into a lower-dimensional space~\cite{mackiewicz1993principal}. 
For each sample, an independent PCA model is trained to compress the features to $\mathbf{h}_i \in \mathbb{R}^{128}$.
Below, we set the embedding dimension $D=128$, which can be yet adjustable to certain tasks.
% \begin{equation}
% \label{eq:visual_feature}
%     \mathbf{h}_i = \text{PCA}(\mathbf{f}_i), \quad \text{where } \mathbf{h}_i \in \mathbb{R}^{128}.
% \end{equation}
% The precise 2D centroid coordinates $\mathbf{c}_i = (x_i, y_i)$ are preserved to define the spatial layout in the subsequent graph construction phase.

\paragraph{\textbf{Cell phenotype clustering.}}
Cell phenotypes are important biological semantics and should be integrated into the modeling. Here, we leverage K-means clustering on the visual features $\mathbf{h}_i$ to learn the morphological prototypes. We cluster on all the cells' visual features of all the training WSI images using the Within-Cluster Sum of Squares (WCSS):
\begin{equation}
    {J} = \sum_{k=1}^{K} \frac{1}{|S_k|} \sum_{i \in S_k} || \mathbf{h}_i - \bar{\mathbf{h}}_k ||^2_2
\label{eq:clustering}
\end{equation}
where $K$ is the number of clusters, $S_k$ the set of cell indexes in the $k$-th cluster, and $\bar{\mathbf{h}}_k$ the cluster centroid. $K$ is typically around 10, representing the number of cell types. The cluster membership serves as a phenotype supervision that will regularize the learning of gene expressions in the later phase.

\subsection{Spatial Enhancement}
\label{subsec:spatial}

Cellular features are also enhanced by their spatial relations, both locally and globally. Locally means capturing the local tissue microenvironment and intercellular interactions, and we model it with locally connected graphs. Globally is to capture cells' long-range relations within the whole tumor, and this is done via a standard Transformer encoder.

\paragraph{\textbf{Local graph construction.}}
Formally, for each patch in a WSI, we construct a spatial graph $\mathcal{G} = ({V}, {E})$.
The node set ${V} = \{v_1, \dots, v_N\}$ represents $N$ segmented cells from a patch. Each node $v_i$ is equipped with a visual feature vector $ \mathbf{h}_i \in \mathbb{R}^{D}$.
The edge set ${E}$ is constructed based on the spatial proximity of cells. We employ $k$-Nearest Neighbors to define connectivity~\cite{guo2003knn}. Specifically, an edge $e_{ij} \in {E}$ is created between node $v_i$ and node $v_j$ if $v_j$ is among the $k$ closest neighbors of $v_i$ w.r.t. their Euclidean distance $||\mathbf{c}_i - \mathbf{c}_j||_2$. A cell's microenvironment is mostly impacted by its surrounding cells, which results in a circle of around eight adjacent cells. So we empirically set $k=8$ to ensure sufficient local context aggregation while balancing computational efficiency.

\paragraph{\textbf{Graph attention embedding.}}
To enrich the cell representations with the local context, we employ a graph attention network (GAT) encoder~\cite{velivckovic2017graph}. Unlike a graph convolutional network that assumes fixed edge weights, GAT allows assigning learnable weights to different neighbors, thus effectively capturing important interactions in the tissue microenvironment.
The GAT encoder consists of $L$ stacked GAT layers. For a given layer $l=0,1,...,L-1$, the input features are denoted as $\mathbf{H}^{(l)} = \{\mathbf{h}_1^{(l)}, \dots, \mathbf{h}_N^{(l)}\}$, where $\mathbf{h}_i^{(0)}=\mathbf{h}_i$. The update rule for node $v_i$ is computed as:

\begin{equation}
    \mathbf{h}_i^{(l+1)} = \text{ReLU}\left( \text{LN} \left( \sum_{j \in \mathcal{N}_i \cup \{i\}} \alpha_{ij}^{(l)} \mathbf{W}^{(l)} \mathbf{h}_j^{(l)} \right) \right)
\end{equation}
where $\mathcal{N}_i$ denotes the set of neighbors for node $v_i$, $\mathbf{W}^{(l)}$ is a learnable weight matrix, and $\text{LN}(\cdot)$ means Layer Normalization. The attention coefficient $\alpha_{ij}^{(l)}$ indicates the importance of neighbor $v_j$ to node $v_i$ and is computed using a multi-head attention mechanism:

\begin{equation}
    \alpha_{ij}^{(l)} = \frac{\exp(\text{LeakyReLU}(\mathbf{a}^\top [\mathbf{W}^{(l)}\mathbf{h}_i^{(l)} \, \oplus \, \mathbf{W}^{(l)}\mathbf{h}_j^{(l)}]))}{\sum_{k \in \mathcal{N}_i \cup \{i\}} \exp(\text{LeakyReLU}(\mathbf{a}^\top [\mathbf{W}^{(l)}\mathbf{h}_i^{(l)} \, \oplus \, \mathbf{W}^{(l)}\mathbf{h}_k^{(l)}]))}
\end{equation}
where $\oplus$ denotes concatenation, and $\mathbf{a}$ is a learnable attention vector. In our implementation, we utilize multi-head attention with 4 heads to stabilize learning, concatenating their outputs followed by a linear projection. The output of the graph encoder is denoted by $\mathbf{H}^{(L)} \in \mathbb{R}^{N \times D}$, where $N$ is the number of cells in a patch, and $D$ is the embedding dimension.

\paragraph{\textbf{Global spatial embedding.}}
To capture the global long-range relations of cells, we use a Transformer to learn this among all the cells within a WSI image.
First, we inject learnable positional encoding into cells. Given the centroid coordinates $(x, y)$ of a cell, we map them to dense vectors using two learnable lookup tables $\mathrm{Emb}_\mathrm{x}, \mathrm{Emb}_\mathrm{y} \in \mathbb{R}^{N \times D}$. The spatial positional encoding $\mathbf{s}_i$ is computed as:
\begin{equation}
    \mathbf{s}_i = \mathrm{Emb}_\mathrm{x}(x_i) \oplus \mathrm{Emb}_\mathrm{y}(y_i)
\end{equation}
These encodings are concatenated to $\mathbf{H}^{(L)}$ and form the input sequence $(\mathbf{H}^{(L)}, \mathbf{S})$ to a Transformer encoder.
The encoder outputs $\mathbf{H}_{\text{cell}} \in \mathbb{R}^{N \times D}$, which represents the final features of cells.
As seen, $\mathbf{H}_{\text{cell}}$ integrates both the visual and spatial characteristics of cells, making it more semantic to subsequent learning.

% \paragraph{\textbf{Cell-Level Prediction Capability.}}
% Since the Transformer output $\mathbf{H}_{\text{cell}}$ already encapsulates high-level biological semantics for each cell, obtaining single-cell gene expression is straightforward. We apply a shared decoding layer $\phi(\cdot)$ (implemented as a lightweight projection head) directly to the Transformer embeddings:
% \begin{equation}
%     \hat{\mathbf{Y}}_{\text{cell}} = \phi(\mathbf{H}_{\text{cell}}) \in \mathbb{R}^{N \times G}
% \end{equation}
% where $\hat{\mathbf{Y}}_{\text{cell}}$ represents the predicted expression of $G$ genes for all $N$ cells. This design ensures that single-cell predictions are driven by the deep contextual features learned by the Transformer, rather than simple local textures.

\subsection{Multiple Instance Learning with Omics}
\label{subsec:mil}

Now, with the cellular features $\mathbf{H}_{\text{cell}}$ extracted, we map them to their gene expressions (or transcriptomics), denoted by $\hat{\mathbf{y}}_{\text{bulk}}$ and $\hat{\mathbf{y}}_{\text{spot}}$. As only bulk or spot-level transcriptome is available for supervision, we adopt multiple instance learning (MIL) to learn the mapping. 
Below, we use spot-level for explanation, and bulk-level is done similarly.

\paragraph{\textbf{Feature-level MIL for spot aggregation.}}
Aggregations of MIL can be done at different stages. One way is to sum up cell-level transcriptomics with equal weights at the label-level. Here we propose a more flexible \textit{feature-level} aggregation strategy using a \textit{gene-specific query attention} mechanism. 
% Instead of aggregating scalar predictions, we aggregate the latent Transformer features $\mathbf{H}_{\text{cell}}$ into robust prototypes.
Concretely, we introduce a learnable query matrix $\mathbf{Q}_\mathrm{gene} \in \mathbb{R}^{G \times D}$ to compute the attention weights between genes and cells:
\begin{equation}
    \mathbf{A} = \text{softmax}\left( \frac{\mathbf{Q}_\mathrm{gene} \mathbf{H}_{\text{cell}}^\top}{\sqrt{D}} \right) \in \mathbb{R}^{G \times N}
\end{equation}
Here, the softmax is applied across the column dimension $N$, ensuring the aggregation is normalized and independent of cell count.
Then, we obtain \textit{gene-specific latent representations} $\mathbf{Z}_{\text{gene}}$, which is an aggregation of differently weighted cellular features, by:
\begin{equation}
    \mathbf{Z}_{\text{gene}} = \mathbf{A} \mathbf{H}_{\text{cell}} \in \mathbb{R}^{G \times D}
\end{equation}
Finally, each gene's representation $\mathbf{z}_g$, a row in $\mathbf{Z}_{\text{gene}}$, can be mapped to the real spot-level expression value by a shared readout function:
\begin{equation}
    \hat{\mathrm{y}}^{\mathrm{spot}}_{g} = \text{Softplus}\left(\mathbf{w}_2^\top \text{ReLU}\left(\mathbf{w}_1^\top \text{LN}(\mathbf{z}_g)\right)\right)
\end{equation}
where $\mathbf{w}_1, \mathbf{w}_2$ are learnable weights. Therefore, we complete the MIL mapping from cellular features $\mathbf{H}_{\text{cell}}$ to the predicted gene expressions $\hat{\mathbf{y}}_{\mathrm{spot}}=\{\hat{\mathrm{y}}^{\mathrm{spot}}_1,...,\hat{\mathrm{y}}^{\mathrm{spot}}_G\}$ at the spot level.

\paragraph{\textbf{Cellular spatial transcriptomics prediction.}}
A side product of our model is that we can generate pseudo cellular ST profiles that are subject to cell phenotype regularization. Recall that the attention weight $\mathbf{A}_{g,i} \in \mathbf{A}$ represents the relative contribution of the $i$-th cell to the total expression of gene $g$ within a spot. We can distribute the spot-level transcriptomic $\hat{\mathbf{y}}_{\text{spot}}$ back to the composed cells $\hat{\mathbf{Y}}_{\text{cell}}$, by:
\begin{equation}
    \hat{\mathrm{y}}_{i,g}^{\text{cell}} = \mathbf{A}_{g,i} \cdot \hat{\mathrm{y}}_{g}^{\text{spot}}
\end{equation}
where $\hat{\mathrm{y}}_{i,g}^{\text{cell}}$ means the $i$-th cell's gene $g$'s predicted expression.
Then a cell $i$'s pseudo RNA-seq profile is denoted by $\hat{\mathbf{y}}_i^{\mathrm{cell}}=\{\hat{\mathrm{y}}^{\mathrm{cell}}_{i,1},...,\hat{\mathrm{y}}^{\mathrm{cell}}_{i,G}\}$

\paragraph{\textbf{Loss function.}}
To train BiTro, we define a composite loss function comprising a gene expression prediction loss and a clustering regularization term. 
Our primary objective is to minimize the discrepancy between the predicted transcriptomics $\hat{\mathbf{y}}_{\mathrm{spot}}$ (or $\hat{\mathbf{y}}_{\mathrm{bulk}}$ for bulk) and the ground truth $\mathbf{y}_{\mathrm{spot}}$ (or $\mathbf{y}_{\mathrm{bulk}}$). We use the Mean Squared Error (MSE) loss:
\begin{equation}
    \mathcal{L}_\mathrm{gene} = \frac{1}{N_\mathrm{s}} \sum_{m=1}^{N_\mathrm{s}} || \hat{\mathbf{y}}_{m}^{\mathrm{spot}} - \mathbf{y}_{m}^{\mathrm{spot}} ||_2^2
\end{equation}
where $N_\mathrm{s}$ is the total number of spots or patches used for training.
Meanwhile, we also want the pseudo cellular transcriptomics $\hat{\mathbf{y}}_{\mathrm{cell}}$ to be consistent with their visual features, for they share the same biological phenotypes. Thus, we introduce a cluster regularization term $\mathcal{L}_\mathrm{cluster}$, which is similar to a standard clustering loss.
Recall that in Eq. (\ref{eq:clustering}), we performed a K-means clustering on cells' visual features. Say we have created $K$ clusters among all the cells segmented from all the training WSI images. Let $S_k$ $(k\in[K])$ be the cell indexes of cluster $k$. Then, the clustering loss defined on the \textit{gene expression} space is:
\begin{equation}
    \mathcal{L}_\mathrm{cluster} = \sum_{k=1}^K \frac{1}{|S_k|} \sum_{i \in S_k} || \hat{\mathbf{y}}_i^{\mathrm{cell}} - \bar{\mathbf{y}}_k ||^2_2
\end{equation}
where $\hat{\mathbf{y}}_i^{\mathrm{cell}}$ is cell $i$'s predicted transcriptomics, and $\bar{\mathbf{y}}_k$ is the mean transcriptomics vector averaged in cluster $k$. We can see that this regularization preserves the consistency of biological semantics.
Finally, we combine the above two losses by a hyperparameter $\lambda>0$:
\begin{equation}
    \mathcal{L}_\mathrm{total} = \mathcal{L}_\mathrm{gene} + \lambda \mathcal{L}_\mathrm{cluster}
\end{equation}
And a standard optimization procedure can be used to work out the model parameters.

\subsection{Bidirectional Transfer Learning with LoRA}

\begin{figure}[htbp]
  \centering
  \includegraphics[width=\linewidth]{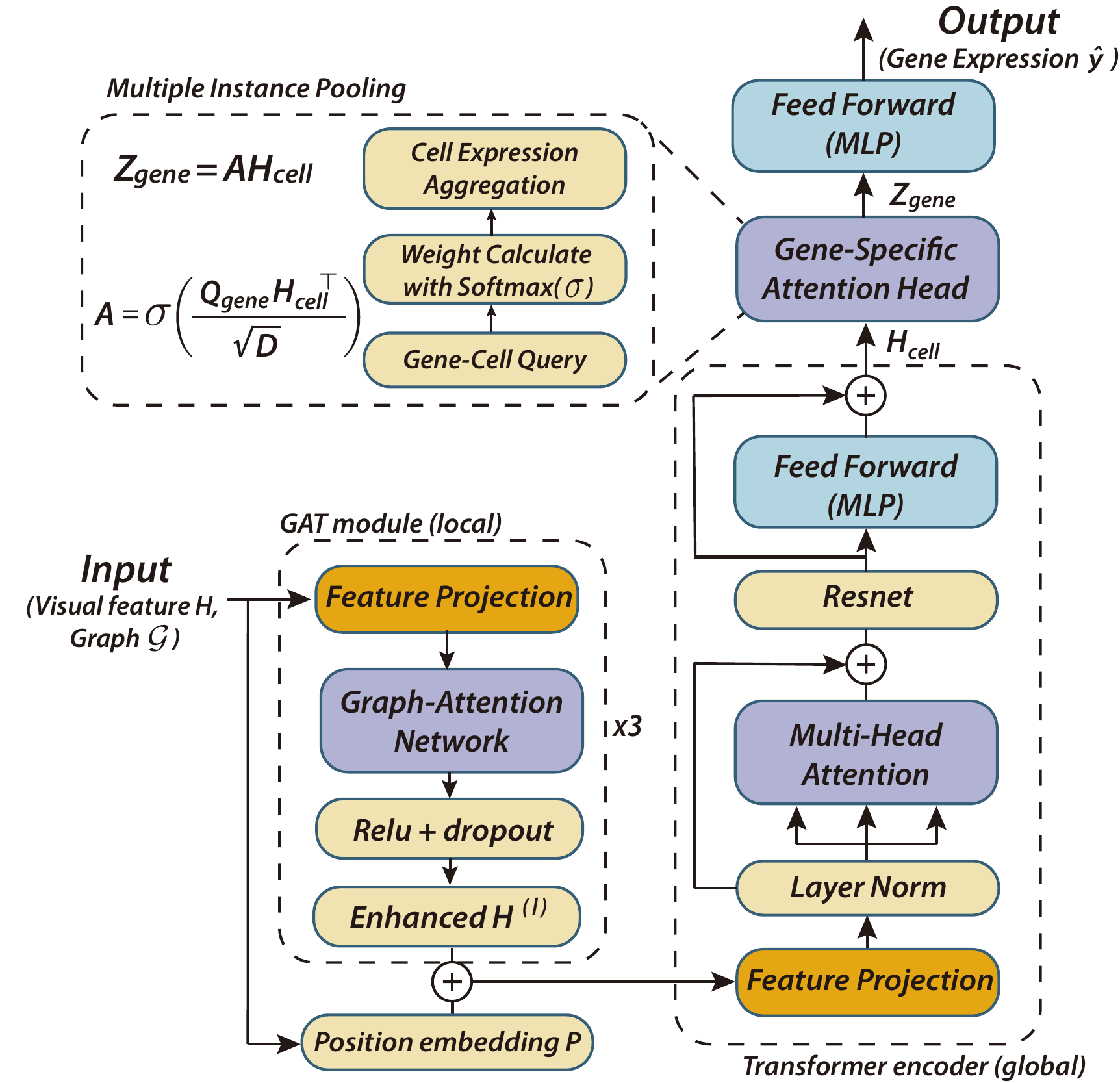}
  \caption{Model architecture. Inputs are cellular visual features $\mathbf{H}$ and their local spatial graph $\mathcal{G}$. The GAT module and the Transformer encoder are to enhance $\mathbf{H}$ with spatial relations both locally and globally. The enhanced features $\mathbf{H}_{\mathrm{cell}}$ are used to predict gene expression $\hat{\mathbf{y}}$ by multiple instance pooling.}
  % (Left) \textbf{LoRA Integration}: The feature projection and attention layers ($W_q, W_v$) are augmented with low-rank decomposition matrices ($A, B$) to facilitate efficient modality adaptation. 
  % (Right) \textbf{Gene-Specific Attention Head}: A learnable gene-cell query mechanism computes attention weights $\alpha$ to aggregate patch features into gene-wise representations $Z_{gene} = \sum_{j=1}^{N}\alpha_{j}H_{cell}$. 
  % The final gene expression is regressed through a ResNet-based Feed-Forward MLP head.}
  \label{fig:network_detail}
\end{figure}

Our base model architecture is summarized in Fig.~\ref{fig:network_detail}, corresponding to Section \ref{subsec:spatial} and \ref{subsec:mil}. This architecture can be used to train one single task, i.e., when one has only bulk+WSI data or ST data. When both kinds of data are available, we can co-use them by transfer learning. This is bidirectional - if the target task is to predict bulk transcriptomics, we can pretrain a model with relevant ST data and finetune the model parameters with the target bulk+WSI data, or vice versa. All the parameters shown in Fig.~\ref{fig:network_detail} are tunable, but some can also be frozen depending on actual tasks.
A little trick of improving finetuning efficiency is to use Low-Rank Adaptation (LoRA). LoRA can project high-dimensional trainable parameters to a low-rank space and then do parameter update.
In our experiments, LoRA is applied to the parameter-heavy components, such as the feature projection layers, the multi-head self-attention modules, and the feed-forward network layers.
% Rather than retraining all parameters of the Transformer and projection layers, we freeze the pre-trained weights $\mathbf{W}_0 \in \mathbb{R}^{d \times k}$ and inject trainable low-rank decomposition matrices $\mathbf{B} \in \mathbb{R}^{d \times r}$ and $\mathbf{A} \in \mathbb{R}^{r \times k}$, where the rank $r \ll \min(d, k)$. The forward pass for a purely linear layer is modified as:
% \begin{equation}
%     \mathbf{h} = \mathbf{W}_0 \mathbf{x} + \frac{\alpha}{r} \mathbf{B}\mathbf{A}\mathbf{x}
% \end{equation}
% where $\alpha$ is a scaling factor. 

\section{Experiments}

In this section, we comprehensively evaluate BiTro's performance across four experiments on five common cancer types. These include bulk-level and spot-level transcriptomics predictions from WSI images, and each task is tested with or without transfer learning. In addition, we also test the pseudo cellular ST profiles on real super-resolution (cellular) spatial transcriptomics data.
Below we present the datasets, experimental setup, and results.

% predicting single-cell (SC) gene expression from whole-slide images (WSI-SC gene prediction), predicting spatial transcriptomics (ST) gene expression from whole-slide images (WSI-ST gene prediction), and predicting bulk transcriptomic expression from whole-slide images (WSI-bulk prediction), as well as three prediction tasks under transfer learning scenarios—namely, the task of predicting single-cell (SC) gene expression from WSI by transferring from a model trained on ST data; WSI-ST gene prediction achieved by post-training on ST data after pre-training on bulk data; and WSI-bulk prediction accomplished by post-training on bulk data following pre-training on ST data, to further validate the model's cross-modal generalization ability across different resolutions.

\subsection{Datasets}
% This study utilized bulk RNA sequencing (Bulk RNA-seq) data, their corresponding hematoxylin-and-eosin (H\&E)-stained whole slide images (WSIs), and spatial transcriptomics (ST) datasets covering different spatial resolutions, thereby providing multi-dimensional and high-quality data support for model training and performance validation.

% \begin{center}
%     \textit{Full data details can be found in Appendix~\ref{appendix:data}.}
% \end{center}

\textbf{Bulk transcriptomics and paired H\&E-stained whole slide images} are retrieved from the TCGA database \cite{weinstein2013cancer}, covering four cancer types: invasive breast carcinoma (BRCA), colon adenocarcinoma (COAD), prostate adenocarcinoma (PRAD), and liver hepatocellular carcinoma (LIHC, corresponding to the HCC type below). Given the large sample size and high heterogeneity of bulk-level data, we only perform one-pass train and test. Each cohort has around 200 to 300 samples, which are randomly split into 4:1. Full data details can be found in Table~\ref{tab:bulk_rna_sample_size}, Appendix~\ref{appendix:data}. For bulk sequencing, TPM (transcripts per million) is the standardized quantification metric measuring gene expression, allowing cross-sample comparison.

\textbf{Spot-level spatial transcriptomics datasets} are retrieved from the Hest-1k platform \cite{jaume2024hest}, covering four cancer types.
BRCA is from the HER2-positive breast tumor dataset \cite{andersson2021spatial}, including 36 samples and 13,620 sequencing spots; 
COAD comprises multiple datasets based on Visium technology \cite{oliveira2024characterization, 10xgenomics_visium_cytassist_colon_cancer, 10xgenomics_human_intestine_cancer, 10xgenomics_human_colorectal_cancer_wta, valdeolivas2024profiling}, including 15 samples and 44,823 spots;
PRAD is from dataset \cite{erickson2022spatially}, including 23 samples and 62,710 spots;
HCC (hepatocellular carcinoma) is from the dataset \cite{giraud2022trem1+} in HEST-Bench, including 2 samples and 4248 spots;
CSCC (cutaneous squamous cell carcinoma) is from the human CSCC dataset \cite{ji2020multimodal}, including 12 samples and 8,671 spots. Note that CSSS does not have an associated bulk dataset. For ST data, we perform leave-one-out cross-validation. 
Details are provided in Table~\ref{tab:low_res_st_counts}, Appendix~\ref{appendix:data}.
UMI (unique molecular identifier) is used to measure gene expression in ST. Since it is a kind of raw molecular count, it can be added and aggregated in spots directly.

\textbf{Super-resolution (subcellular) spatial transcriptomics data} are also retrieved from Hest-1k, covering four cancer types.
As said, such public data are few, so each cancer has only one sample. 
BRCA is from a Xenium breast cancer dataset generated by 10x Genomics, and Replicate 1 is used \cite{janesick2023high},  containing 167,780 cells; COAD is from a FFPE human colorectal cancer dataset generated by 10x Genomics with human immuno-oncology profiling panel and custom add-on \cite{10xgenomics_ffpe_human_colorectal_cancer_immuno_oncology}, containing 388,175 cells; PRAD is from a Xenium Prime FFPE human prostate dataset generated by 10x Genomics \cite{10xgenomics_xenium_prime_ffpe_human_prostate}, containing 193,000 cells; LIHC is a from human liver dataset generated by 10x Genomics with Xenium human multi-tissue and cancer panel \cite{10xgenomics_human_liver_xenium_multi_tissue_cancer}, containing 162,628 cells. Each sample is spatially divided into five pieces, and five-fold cross-validation is performed. Details are provided in Table~\ref{tab:super_res_st_counts}, Appendix~\ref{appendix:data}. Transcript count in Xenium technology has the same characteristic as UMI, which can be added or aggregated in cells or spots directly.

\textbf{Gene selection} of above datasets can be found in Appendix~\ref{appendix:data}.

\subsection{Evaluation Metrics}
% We conducted gene prediction experiments across three dimensions—cell, spatial transcriptomics, and bulk. And we designated distinct evaluation methods for each dimension. 
% Given that none of the previous bulk-level models calculated gene-level correlation, we did not compute this metric.
% As for the evaluation method for spatial transcriptomics, confusion regarding the PCC metric exists in previously published studies in spatial transcriptomics prediction, for example, in the paper introducing Histogene, gene Pearson was adopted as the evaluation metric. However, in the paper introducing M2OST, overall Pearson was adopted as the evaluation metric. We therefore calculated both metrics in our experiments.
% For cell-level gene expression prediction, the evaluation was performed in two ways: first, calculating the overall correlation between the predicted and original gene expression profiles for each individual cell; second, computing the mean value of correlation coefficients across different genes across all cells. 

To quantify gene expression prediction accuracy, we adopt two metrics: Pearson Correlation Coefficient (PCC) and Jensen-Shannon (JS). 
PCC is commonly used to measure the linear correlation between a model's predicted values and the true values. 
Two versions of PCC have been previously used. 
$\text{PCC}_{\text{overall}}$ is to measure the correlation between all genes' predicted vs. true values in a unit (i.e., in a bulk or a spot), given by:
\begin{equation}
    \text{PCC}_{\text{overall}} = \frac{\sum_{g=1}^{G} (\mathrm{y}_g - \bar{\mathrm{y}})(\hat{\mathrm{y}}_g - \bar{\hat{\mathrm{y}}})}{\sqrt{\sum_{g=1}^{G} (\mathrm{y}_g - \bar{\mathrm{y}})^2} \cdot \sqrt{\sum_{g=1}^{G} (\hat{\mathrm{y}}_g - \bar{\hat{\mathrm{y}}})^2}}
\end{equation}
And $\text{PCC}_{\text{gene}}$ is to measure individual genes' predicted vs. true values across all $S$ units, given by:
\begin{equation}
    \text{PCC}_{\text{gene}}(g) = \frac{\sum_{s=1}^{S} (\mathrm{y}_{s,g} - \bar{\mathrm{y}}_g)(\hat{\mathrm{y}}_{s,g} - \bar{\hat{\mathrm{y}}}_g)}{\sqrt{\sum_{s=1}^{S} (\mathrm{y}_{s,g} - \bar{\mathrm{y}}_g)^2} \cdot \sqrt{\sum_{s=1}^{S} (\hat{\mathrm{y}}_{s,g} - \bar{\hat{\mathrm{y}}}_g)^2}}
\end{equation}
Our results report averaged $\text{PCC}_{\text{overall}}$ across all units and averaged $\text{PCC}_{\text{gene}}$ across all genes.
The second metric, JS divergence, measures the distributional difference between two sets of gene expressions. This is particularly used for measuring bulk transcriptomics, as each bulk's expression is commonly normalized in TPM (transcripts per million). Say we have a bulk transcriptomics vector normalized in a probability distribution $\mathbf{y}_{\text{bulk}}$ and its prediction $\mathbf{\hat{y}}_{\text{bulk}}$. Their JS divergence is defined as:
\begin{equation}
\text{JS}(\mathbf{y}_{\text{bulk}} \parallel \mathbf{\hat{y}}_{\text{bulk}}) = \frac{1}{2} \text{KL}(\mathbf{y}_{\text{bulk}} \parallel \mathbf{\hat{y}}_{\text{bulk}}) + \frac{1}{2} \text{KL}(\mathbf{y}_{\text{bulk}} \parallel \mathbf{\hat{y}}_{\text{bulk}})
\end{equation}
where $\text{KL}$ is the Kullback-Leibler divergence.

\subsection{Results}

\begin{table*}[t]
  \caption{Performance comparison on bulk transcriptomics prediction. Best results are bolded, and the second best underlined.}
  \label{tab:bulk_final}
  \centering
  \small
  \setlength{\tabcolsep}{10pt} 
  
  \begin{tabular*}{\textwidth}{@{\extracolsep{\fill}} l cc cc cc cc}
    \toprule
    \textbf{Method} & \multicolumn{2}{c}{\textbf{BRCA}} & \multicolumn{2}{c}{\textbf{COAD}} & \multicolumn{2}{c}{\textbf{PRAD}} & \multicolumn{2}{c}{\textbf{LIHC (HCC)}} \\
    
    \cmidrule(lr){2-3} \cmidrule(lr){4-5} \cmidrule(lr){6-7} \cmidrule(lr){8-9}

    & $\text{PCC}_\text{overall} \uparrow$ & \hspace{-12pt}JS $\downarrow$ & $\text{PCC}_\text{overall} \uparrow$ & \hspace{-12pt}JS $\downarrow$ & $\text{PCC}_\text{overall} \uparrow$ & \hspace{-12pt}JS $\downarrow$ & $\text{PCC}_\text{overall} \uparrow$ & \hspace{-12pt}JS $\downarrow$ \\
    \midrule
    Original  & 0.502{\scriptsize $\pm$0.198} & \hspace{-15pt}0.181{\scriptsize $\pm$0.046} & 0.869{\scriptsize $\pm$0.015} & \hspace{-15pt}0.040{\scriptsize $\pm$0.015} & 0.659{\scriptsize $\pm$0.148} & \hspace{-15pt}0.092{\scriptsize $\pm$0.034} & 0.662{\scriptsize $\pm$0.248} & \hspace{-15pt}0.214{\scriptsize $\pm$0.076} \\
    UNI+MLP  & 0.652{\scriptsize $\pm$0.275} & \hspace{-15pt}0.132{\scriptsize $\pm$0.059} & 0.920{\scriptsize $\pm$0.098} & \hspace{-15pt}0.021{\scriptsize $\pm$0.015} & 0.806{\scriptsize $\pm$0.127} & \hspace{-15pt}0.053{\scriptsize $\pm$0.021} & 0.735{\scriptsize $\pm$0.177} & \hspace{-15pt}0.174{\scriptsize $\pm$0.062} \\
    HE2RNA     & 0.710{\scriptsize $\pm$0.032} & \hspace{-15pt}\textbf{0.086{\scriptsize $\pm$0.002}} & 0.911{\scriptsize $\pm$0.005} & \hspace{-15pt}0.024{\scriptsize $\pm$0.002} & \underline{0.830{\scriptsize $\pm$0.157}} & \hspace{-15pt}\underline{0.051{\scriptsize $\pm$0.005}} & \underline{0.827{\scriptsize $\pm$0.234}} & \hspace{-15pt}\underline{0.119{\scriptsize $\pm$0.084}} \\
    SEQUOIA    & \underline{0.738{\scriptsize $\pm$0.199}} & \hspace{-15pt}0.094{\scriptsize $\pm$0.059} & 0.932{\scriptsize $\pm$0.043} & \hspace{-15pt}\underline{0.020{\scriptsize $\pm$0.022}} & 0.820{\scriptsize $\pm$0.068} & \hspace{-15pt}0.056{\scriptsize $\pm$0.019} & 0.738{\scriptsize $\pm$0.109} & \hspace{-15pt}0.149{\scriptsize $\pm$0.014} \\
    \midrule
    \textbf{BiTro (w/o trans.)} & 0.711{\scriptsize $\pm$0.174} & \hspace{-15pt}0.107{\scriptsize $\pm$0.081} & \underline{0.933{\scriptsize $\pm$0.007}} & \hspace{-15pt}0.021{\scriptsize $\pm$0.009} & 0.814{\scriptsize $\pm$0.127} & \hspace{-15pt}0.050{\scriptsize $\pm$0.022} & 0.811{\scriptsize $\pm$0.263} & \hspace{-15pt}0.139{\scriptsize $\pm$0.093} \\
    \textbf{BiTro (trans)} & \textbf{0.743{\scriptsize $\pm$0.186}} & \hspace{-15pt}\underline{0.091{\scriptsize $\pm$0.058}} & \textbf{0.946{\scriptsize $\pm$0.007}} & \textbf{\hspace{-15pt}0.019{\scriptsize $\pm$0.014}} & \textbf{0.831{\scriptsize $\pm$0.112}} & \textbf{\hspace{-15pt}0.043{\scriptsize $\pm$0.012}} & \textbf{0.844{\scriptsize $\pm$0.194}} & \textbf{\hspace{-15pt}0.117{\scriptsize $\pm$0.102}} \\
    \bottomrule
  \end{tabular*}
\end{table*}

\paragraph{\textbf{Bulk transcriptomics prediction.}}
Table \ref{tab:bulk_final} presents the performance comparison between our BiTro model (with or without transfer) vs. other models on bulk transcriptomics prediction. The results are averaged among all the test cases.
The comparative methods are Original (arbitrarily two bulk transcriptomics vectors' correlation and similarity), UNI+MLP (a variation of SEQUOIA~\cite{pizurica2024digital}), HE2RNA~\cite{schmauch2020deep}, and SEQUOIA (using UNI+ViT, a SOTA method). Their implementation details can be found in Appendix~\ref{appendix:implementation}. The purpose of introducing Original is to stress that our results are no worse than random comparisons between any two real cases.
Across four cancer types, four comparative models, and two evaluation metrics, BiTro with transfer achieves almost all the best results.
In the transfer learning mode, we first pretrain the model with the associated spot-level ST data and then finetune the model parameters on the bulk dataset, which ultimately yields promising performance. This validates the effectiveness of BiTro with transfer in bulk gene expression prediction.
In the biology context, the position information of ST data enables the model to capture spatial information better. In the machine learning context, pretraining on similar tasks with stronger supervision allows the model to converge to a relatively better parameter space in advance.
We also want to note that even without transfer, Bitro can also achieve very competitive performance compared to other models.
% On the BRCA dataset, our model achieves the highest Pearson of 0.743 (±0.186) and lowest JS divergence of 0.091 (±0.058). 
% On the COAD dataset, our model achieves the highest Pearson of 0.946 (±0.007) and the second lowest JS divergence of 0.019 (±0.014).
% On the PRAD dataset, our model achieves a Pearson correlation coefficient of 0.831 (±0.127) and JS divergence of 0.050 (±0.022), slightly outperforming HE2RNA (0.830 and 0.049), showcasing stable and superior performance. 

\begin{table*}[t]
  \caption{Performance comparison on spatial transcriptomics prediction. Best results are bolded, and the second best underlined.}
  \label{tab:st_comparison}
  \centering
  \small
  \setlength{\tabcolsep}{2.5pt} 
  \begin{tabular*}{\textwidth}{@{\extracolsep{\fill}}lcccccccccc}
    \toprule
    \textbf{Model} & \multicolumn{2}{c}{\textbf{BRCA}} & \multicolumn{2}{c}{\textbf{COAD}} & \multicolumn{2}{c}{\textbf{PRAD}} & \multicolumn{2}{c}{\textbf{HCC}} & \multicolumn{2}{c}{\textbf{CSCC}} \\
    \cmidrule(lr){2-3} \cmidrule(lr){4-5} \cmidrule(lr){6-7} \cmidrule(lr){8-9} \cmidrule(lr){10-11}
    & $\text{PCC}_\text{overall} \uparrow$ & $\text{PCC}_\text{gene} \uparrow$ & $\text{PCC}_\text{overall} \uparrow$ & $\text{PCC}_\text{gene} \uparrow$  & $\text{PCC}_\text{overall} \uparrow$ & $\text{PCC}_\text{gene} \uparrow$  & $\text{PCC}_\text{overall} \uparrow$ & $\text{PCC}_\text{gene} \uparrow$  & $\text{PCC}_\text{overall} \uparrow$ & $\text{PCC}_\text{gene} \uparrow$  \\
    \midrule
    
    HisToGene  & 0.541{\scriptsize $\pm$0.079} & 0.114{\scriptsize $\pm$0.079} & 0.515{\scriptsize $\pm$0.161} & 0.102{\scriptsize $\pm$0.062} & 0.449{\scriptsize $\pm$0.120} & 0.110{\scriptsize $\pm$0.087} & 0.282{\scriptsize $\pm$0.020} & 0.072{\scriptsize $\pm$0.005} & 0.560{\scriptsize $\pm$0.049} & 0.157{\scriptsize $\pm$0.086} \\
    Hist2ST   & 0.574{\scriptsize $\pm$0.090} & 0.141{\scriptsize $\pm$0.084} & 
    \textbf{0.621{\scriptsize $\pm$0.080} } & \underline{0.133{\scriptsize $\pm$0.132}} &
    0.438{\scriptsize $\pm$0.160} & 0.071{\scriptsize $\pm$0.057} & 0.274{\scriptsize $\pm$0.013} & -0.001{\scriptsize $\pm$0.005} & \underline{0.610{\scriptsize $\pm$0.109}} & 0.131{\scriptsize $\pm$0.061} \\
    THItoGene & 0.558{\scriptsize $\pm$0.099} & 0.094{\scriptsize $\pm$0.089} & 0.614{\scriptsize $\pm$0.076} & 0.075{\scriptsize $\pm$0.081} & 0.484{\scriptsize $\pm$0.104} & 0.134{\scriptsize $\pm$0.106} & 0.247{\scriptsize $\pm$0.051} & 0.005{\scriptsize $\pm$0.003} & 0.599{\scriptsize $\pm$0.039} & \underline{0.241{\scriptsize $\pm$0.044}} \\
    M2OST     & 0.558{\scriptsize $\pm$0.660} & 0.117{\scriptsize $\pm$0.081} & 0.514{\scriptsize $\pm$0.057} & 0.122{\scriptsize $\pm$0.117} & \underline{0.515{\scriptsize $\pm$0.094}} & \textbf{0.256{\scriptsize $\pm$0.070}} & 0.124{\scriptsize $\pm$0.002} & 0.102{\scriptsize $\pm$0.030} & 0.450{\scriptsize $\pm$0.084} & 0.142{\scriptsize $\pm$0.057} \\
    \midrule
    \textbf{BiTro (w/o trans.)} & \underline{0.587{\scriptsize $\pm$0.101}} & \underline{0.154{\scriptsize $\pm$0.136}} & 0.565{\scriptsize $\pm$0.066} & 0.103{\scriptsize $\pm$0.092} & 0.502{\scriptsize $\pm$0.129} & 0.186{\scriptsize $\pm$0.075} & \underline{0.341{\scriptsize $\pm$0.012}} & \underline{0.128{\scriptsize $\pm$0.025}} &
    \textbf{0.667{\scriptsize $\pm$0.041}} & \textbf{0.429{\scriptsize $\pm$0.154}} \\
    \textbf{BiTro (trans.)} & \textbf{0.598{\scriptsize $\pm$0.100}} & \textbf{0.173{\scriptsize $\pm$0.136}} &\underline{0.615{\scriptsize $\pm$0.064}} & \textbf{0.137{\scriptsize $\pm$0.101}} & \textbf{0.521{\scriptsize $\pm$0.132}} & \underline{0.214{\scriptsize $\pm$0.116}} & \textbf{0.342{\scriptsize $\pm$0.012}} & \textbf{0.144{\scriptsize $\pm$0.011}} &
    --- & --- \\
    \bottomrule
  \end{tabular*}
\end{table*}
\begin{table*}[t]
  \caption{Performance comparison on cellular transcriptomics prediction. Best results are bolded, and the second best underlined.}
  \label{tab:sc_comparison}
  \centering
  \small
  \setlength{\tabcolsep}{4pt} 
  \begin{tabular*}{\textwidth}{@{\extracolsep{\fill}}l cc cc cc cc}
    \toprule
    \textbf{Model} & \multicolumn{2}{c}{\textbf{BRCA}} & \multicolumn{2}{c}{\textbf{COAD}} & \multicolumn{2}{c}{\textbf{PRAD}} & \multicolumn{2}{c}{\textbf{HCC}} \\
    \cmidrule(lr){2-3} \cmidrule(lr){4-5} \cmidrule(lr){6-7} \cmidrule(lr){8-9}
    & $\text{PCC}_\text{overall} \uparrow$ & $\text{PCC}_\text{gene} \uparrow$ & $\text{PCC}_\text{overall} \uparrow$ & $\text{PCC}_\text{gene} \uparrow$ & $\text{PCC}_\text{overall} \uparrow$ & $\text{PCC}_\text{gene} \uparrow$ & $\text{PCC}_\text{overall} \uparrow$ & $\text{PCC}_\text{gene} \uparrow$ \\
    \midrule

    GHIST & 0.499{\scriptsize$\pm$0.024} & \textbf{0.348{\scriptsize$\pm$0.038}} & 0.450{\scriptsize$\pm$0.092} & \textbf{0.209{\scriptsize$\pm$0.056}} & 0.241{\scriptsize$\pm$0.010} & \textbf{0.141{\scriptsize$\pm$0.009}} & 0.509{\scriptsize$\pm$0.070} & \textbf{0.204{\scriptsize$\pm$0.108}} \\
    
    iStar & \textbf{0.562{\scriptsize$\pm$0.037}} & \underline{0.223{\scriptsize$\pm$0.018}} & \textbf{0.639{\scriptsize$\pm$0.024}} & \underline{0.153{\scriptsize$\pm$0.012}} & 0.289{\scriptsize$\pm$0.020} & \underline{0.094{\scriptsize$\pm$0.016}} & \textbf{0.699{\scriptsize$\pm$0.033}} & \underline{0.166{\scriptsize$\pm$0.016}} \\

    \midrule
    \textbf{BiTro (w/o trans.)} & 
    0.512{\scriptsize$\pm$0.026} & 0.089{\scriptsize$\pm$0.034} & 0.513{\scriptsize$\pm$0.041} & 0.084{\scriptsize$\pm$0.042} & \underline{0.328{\scriptsize$\pm$0.031}} & 0.051{\scriptsize$\pm$0.021} & \underline{0.587{\scriptsize$\pm$0.042}} & 0.084{\scriptsize$\pm$0.017} \\
    
    \textbf{BiTro (trans)} & \underline{0.544{\scriptsize$\pm$0.041}} & 0.135{\scriptsize$\pm$0.024} & \underline{0.533{\scriptsize$\pm$0.047}} & 0.103{\scriptsize$\pm$0.041} & \textbf{0.342{\scriptsize$\pm$0.019}} & 0.082{\scriptsize$\pm$0.019} & 0.542{\scriptsize$\pm$0.036} & 0.113{\scriptsize$\pm$0.012} \\
    \bottomrule
  \end{tabular*}
\end{table*}

\begin{figure*}[t] 
    \centering
    \includegraphics[width=\textwidth]{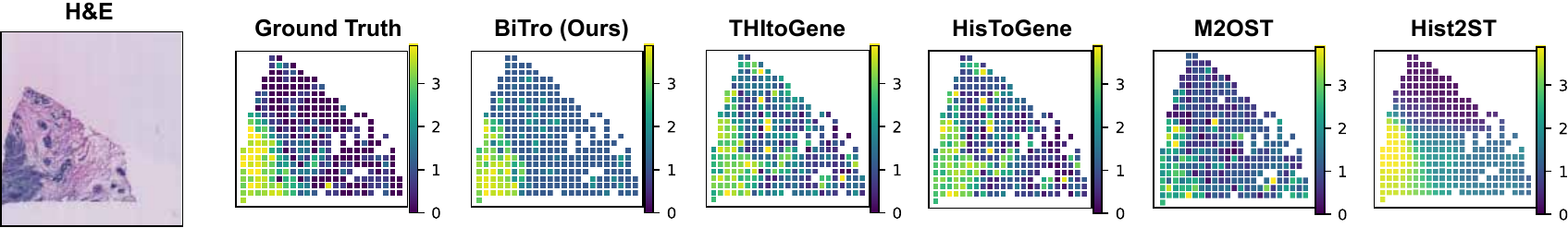} 
    \caption{Gene expression visualization on the SPA148 sample of BRCA for gene in log1p space.}
    \Description{A visualization showing the gene expression heatmaps for GNAS on the SPA148 dataset.}    
    \label{fig:st_illustration}
\end{figure*}

\paragraph{\textbf{Spatial transcriptomics prediction.}}
Table \ref{tab:st_comparison} presents the performance comparison on spatial transcriptomics prediction. Results are averaged by leave-one-out cross-validation. The comparative methods are HisToGene~\cite{pang2021leveraging}, Hist2ST \cite{zeng2022spatial}, THItoGene \cite{jia2023thitogene}, and M2OST~\cite{wang2025m2ost}, representing the SOTA standard. Our BiTro model, both with or without transfer learning, achieves even better performance compared to bulk transcriptomics prediction. On the BRCA, HCC, and CSCC datasets, we BiTro with transfer and without transfer win the top two places, and on COAD and PRAD, Bitro also shows superior performance. 
In the transfer learning mode, we pretrain the model using the associated bulk data and then finetune it with ST data, which ultimately yields promising performance. In the biology context, bulk sequencing provides greater sequencing depth to complement the shallow sequencing depth of ST. In the machine learning context, pretraining on similar tasks allows the model to converge to a relatively better parameter space in advance.
Regarding the two versions of Pearson correlations, they are not always positively correlated but reflect different model abilities.
High overall Pearson indicates that models can accurately reconstruct the overall distribution structure of gene expression profiles across different units; and high gene Pearson suggests that models are good at predicting certain gene expression trends across different units.
Furthermore, we illustrate a sample ST prediction in Fig.~\ref{fig:st_illustration} that is on the SPA148 sample of BRCA for gene GNAS.
The above results demonstrate the strong generalization capability of our model across diverse spatial transcriptomics datasets.

% In the context of predicting gene expression from imaging data, we focus not only on the gene distribution within individual spots or cells but also on the inter-spot and inter-cell variations in the expression of the same gene. So both aspects are equally important.
% However in the context of weakly supervised learning, the gene correlation is more capable of capturing the inter-sample variations. In contrast, a model can yield a high sample correlation merely by learning the global trends while neglecting such variations, which is not the performance we intend to measure with this metric. For this reason, the gene-level correlation is of greater importance. In this context, our model achieved the best performance metrics across four out of the five cancer types investigated.

%迁移解释

\paragraph{\textbf{Cellular spatial transcriptomics prediction.}}

\begin{figure*}[t]
    \centering
    \includegraphics[width=\textwidth]{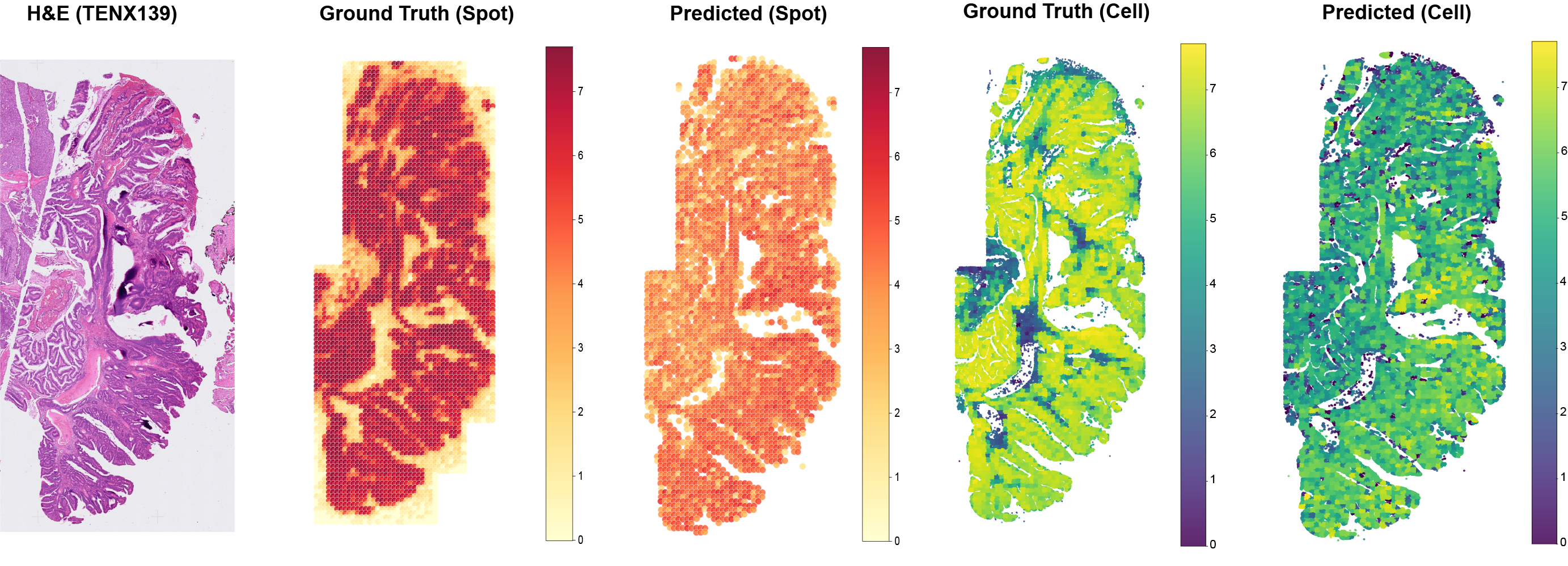}
    \caption{Visualization of BiTro's inference on super-resolved spatial gene expression for gene EPCAM in the COAD sample TENX139 at both spot and cellular levels in the log1p space.}
    \label{fig:scell}
\end{figure*}

As introduced, our BiTro model can generate pseudo cellular ST profiles as intermediate variables, which are not the final output. Yet, we also present our prediction results in Table \ref{tab:sc_comparison} in comparison with iStar \cite{zhang2024inferring} and GHIST \cite{fu2025spatial}, which are specialized in cellular ST prediction.
iStar is designed to improve the resolution of spot-level spatial transcriptomics, and GHIST is a model trained with super-resolution ST data and used for cellular ST prediction.
We conduct experiments on four Xenium datasets of four types of cancer. The experiment setup is the same as iStar's, that is, train our model with pseudo spot-level ST data that are made up of cellular ST data, and then test it on real cellular ST data. We perform five-fold cross-validation where the four folds are used to make pseudo ST data and the rest for test.
According to the results, GHIST is leading in gene Pearson, iStar is leading in overall Pearson, and ours follows with comparable performance (see Fig.~\ref{fig:scell} for a prediction visualization). GHIST achieves the best with no doubt since it is trained with the single-cell level supervision.
iStar, however, also uses pseudo-ST supervision but aims specifically for cell-level spatial transcriptomics prediction, thus also achieving leading performance.
Nevertheless, we also want to stress that our framework is cell-centric, offering room for integrating single-cell RNA sequencing (scRNA-seq) data. We will explore incorporating scRNA-seq as an auxiliary modality to provide explicit, fine-grained supervision for our cell-level feature encoders, thereby moving beyond weak supervision to further refine cellular expression predictions.

\subsection{Ablation Study}

To validate the effectiveness of BiTro's key components, we conducted comprehensive ablation studies across both spatial transcriptomics and bulk transcriptomics tasks. The quantitative results, which are detailed in Appendix \ref{appendix:ablation}, yield several primary insights into our framework's design.

First, the integration of bidirectional transfer learning consistently boosts performance compared to training from scratch. This confirms that distilling knowledge from a source modality—whether the high sequencing depth of bulk data or the spatial resolution of ST data—enhances the model's generalization capability in the target domain. Second, we observe that the impact of phenotype clustering is highly dependent on data granularity. For coarse-grained bulk transcriptomics, clustering provides essential fine-grained regularization to compensate for sparse supervision; however, for spatially resolved ST data, this additional constraint is less critical. Finally, variance normalization proves vital for stabilizing the learning process. By mitigating the dominance of high-variance genes, it leads to universal improvements in Gene Pearson correlations, ensuring the model effectively captures a diverse and balanced range of gene expression patterns.

\section{Limitations and Ethical Considerations}

Despite the promising results, our BiTro framework has a few limitations. First, the computational complexity of the cell-level feature extraction for massive Whole Slide Images (WSIs) is high, making the pipeline relatively time-consuming. Second, while BiTro excels in bulk and spot-level predictions, its single-cell resolution accuracy still has room for improvement. Since the model is trained under weak supervision from aggregated signals, it may not fully capture the precise patterns inherent in true single-cell sequencing data, which should be solved by introducing scRNA-seq data for supervision or some other data mining techniques. 

The datasets involved in our experiments are derived from three public sources: \textbf{TCGA} (for bulk transcriptomics and WSIs), \textbf{HEST-1k} (for spatial transcriptomics), and \textbf{10x Genomics} public repositories (for Xenium super-resolution data).
All these datasets are publicly available and have been de-identified by the original providers to ensure patient privacy.
Our use of the abovementioned datasets is strictly under their regulations and policies. Neither do we have any copyright issues with these datasets.

\section{Conclusions and Future Work}

We propose and develop BiTro, a unified and bidirectional transfer learning framework designed to map pathological images to transcriptomic expression using multiple granularities of data. By grounding our architecture on cell-level visual and spatial features, BiTro effectively bridges the gap between coarse-grained bulk transcriptomics and fine-grained spatial transcriptomics. A key contribution of our study is the validation that bidirectional transfer learning can indeed enhance prediction performance compared to single-task training; the high sequencing depth of bulk data complements the sparsity of spatial data, while the explicit spatial resolution of ST data informs the deconvolution of bulk profiles. Extensive experiments across five cancer cohorts validate that BiTro achieves state-of-the-art performance in both bulk and spot-level transcriptomics prediction tasks. Furthermore, despite being trained under weak supervision, the framework demonstrates the potential of inferring interpretable single-cell gene expression profiles.

Looking ahead, our framework can be extended to two promising directions. First, the cell-centric architecture of BiTro offers room for integrating single-cell RNA sequencing (scRNA-seq) data. Future work will explore incorporating scRNA-seq as an auxiliary modality to provide explicit, fine-grained supervision for our cell-level feature encoders, thereby moving beyond weak supervision to further refine cellular expression predictions. Second, having successfully established the transfer logic between spatial transcriptomics and bulk transcriptomics, we aim to extend this bridge to downstream clinical tasks, for example, cancer survival analysis.
We plan to leverage the rich survival and clinical outcome data available in the TCGA database—which is paired with bulk data—enhanced by the high-fidelity spatial context transferred from ST models. 
% This cross-modal synergy holds significant potential for improving the accuracy of patient prognosis and survival prediction models in precision oncology.

\section{GenAI Disclosure}

The whole content of this paper, including texts, figures, and tables, are not generated by any generative AI technology.

\bibliographystyle{ACM-Reference-Format}
\bibliography{sample-base}


\appendix
\section{Data Details}
Table \ref{tab:bulk_rna_sample_size} to \ref{tab:super_res_st_counts} summarize the dataset details of bulk+WSI cases, spot-level ST data, and cellular ST data, respectively. 
Regarding gene selection, for the BRCA and CSCC spatial transcriptomics (ST) datasets, we selected the same highly variable genes (HVGs) as utilized in existing models such as Hist2ST. For the PRAD, COAD, and HCC datasets, we identified varying numbers of HVGs for model training according to the selection criteria detailed in Appendix B. To facilitate subsequent transfer learning, the bulk transcriptomics models were trained on the intersection of the original gene set and the HVGs identified from the ST data. For inference on single-cell super-resolution data, we utilized the intersection of genes available in the Xenium dataset and the full gene set from the ST data. Notably, for the PRAD dataset, due to an excessively large overlapping gene set, we further refined our selection to the top 200 HVGs within that intersection.
\label{appendix:data}

\begin{table}[!htbp]
  \caption{Dataset statistics of bulk transcriptomics and WSI cases.}
  \label{tab:bulk_rna_sample_size}
  \centering
  \small
  \setlength{\tabcolsep}{2pt}
  \begin{tabular*}{\linewidth}{@{\extracolsep{\fill}}lccccc}
    \toprule
    \textbf{Set} & \textbf{BRCA} & \textbf{COAD} & \textbf{LIHC (HCC)} & \textbf{PRAD} \\
    \midrule
    Training    & 200 & 246 & 104 & 152 \\
    Test        & 50 & 59 & 27 & 38 \\
    Total       & 250 & 305 & 131 & 190 \\
    Genes       & 734 & 388 & 185 & 175 \\
    \bottomrule
    \multicolumn{6}{l}{\footnotesize Note: CSCC has no Bulk transcriptomics data in this study.}
  \end{tabular*}
\end{table}

\begin{table}[!htbp]
  \caption{Dataset statistics of spot-level ST datasets.}
  \label{tab:low_res_st_counts}
  \centering
  \small
  \setlength{\tabcolsep}{2pt}
  \begin{tabular*}{\linewidth}{@{\extracolsep{\fill}}lccccc}
    \toprule
    \textbf{Metric} & \textbf{BRCA} & \textbf{COAD} & \textbf{CSCC} & \textbf{LIHC (HCC)} & \textbf{PRAD}   \\
    \midrule
    Technology    & ST   &  Visium &  ST  & Visium & Visium \\
    Samples       & 36  & 15 & 12 & 2 & 23 \\
    Total Spots   & 13,620 & 44,823 & 8,671 & 4248 & 62,710 \\
    Min Spots/Sample & 178 & 328 & 462 & 1987 & 1,418 \\
    Genes         & 785 & 388 & 469 & 200 & 200 \\
    \bottomrule

  \end{tabular*}
\end{table}

\begin{table}[!htbp]
  \caption{Dataset statistics of super-resolution (cellular) ST datasets.}
  \label{tab:super_res_st_counts}
  \centering
  \small
  \setlength{\tabcolsep}{2pt}
  \begin{tabular*}{\linewidth}{@{\extracolsep{\fill}}lcccc}
    \toprule
    \textbf{Metric} & \textbf{BRCA} & \textbf{COAD} & \textbf{LIHC (HCC)} & \textbf{PRAD} \\
    \midrule
    Technology    & Xenium   &  Xenium &  Xenium  & Xenium \\
    ID       &  NCBI785  & TENX139 & TENX120 & TENX157\\
    Total cells   & 167,780 & 388,175 & 162,628 & 193,000 \\
    Median transcripts/cell & 166.0 & 78.0 & 114.0 & 144.0 \\
    Transcripts per \si{100 \um\squared} & 106.0 & 182.8 & 85.0 & 371.7 \\
    Genes   & 278 & 465 & 474 & 300 \\
    \bottomrule
  \end{tabular*}
\end{table}

\section{Data Preprocessing Details}
\label{appendix:preprocess}

\paragraph{\textbf{WSI stain normalization.}}
Histological images often suffer from batch effects due to variations in staining protocols and scanner calibrations. To standardize the visual appearance while preserving tissue morphology, we employ the structure-preserving Vahadane color normalization method.~\cite{vahadane2016structure}.   
Since pixel intensity in the RGB space is non-linear with respect to stain amount, we first transform the RGB image into the \textit{Optical Density} (OD) space, denoted by $\mathbf{V}$, based on the Beer-Lambert law.
We assume the image is composed of two primary stains: Hematoxylin (nuclei) and Eosin (cytoplasm). Using Sparse Non-negative Matrix Factorization (SNMF), we decompose $V$ into a stain color matrix $\mathbf{W}_\mathrm{stain}$ and a stain density map $\mathbf{H}_\mathrm{stain}$:
\begin{equation}
    \mathbf{V} \approx \mathbf{W}_\mathrm{stain}\mathbf{H}_\mathrm{stain},
\end{equation}
Here, columns of $\mathbf{W}_\mathrm{stain} \in \mathbb{R}^{3 \times 2}$ represent the specific RGB absorption vectors for Hematoxylin and Eosin, while rows of $\mathbf{H}_\mathrm{stain} \in \mathbb{R}^{2 \times N}$ represent the concentration of each stain at every pixel.
To align the color distribution, we estimate the specific concentration map $\mathbf{H}_\mathrm{src}$ of the source image and project it onto a pre-defined reference stain matrix $\mathbf{W}_\mathrm{ref}$:
\begin{equation}
    \mathbf{V}_\mathrm{norm} = \mathbf{W}_\mathrm{ref} \mathbf{H}_\mathrm{src}.
\end{equation}
Essentially, this process retains the original tissue structure (encoded in $H_\mathrm{src}$) but "re-paints" it using the standard colors defined by $W_\mathrm{ref}$. Finally, $V_\mathrm{norm}$ is transformed back to RGB space.

\paragraph{\textbf{Cell segmentation.}}
Precise spatial positioning is fundamental for feature extraction and modeling intercellular correlations. 
To obtain this information, for TCGA WSIs\cite{weinstein2013cancer}, we employ HoverNet to derive the 2D centroid coordinates $\mathbf{c}_i = (x_i, y_i)$ for cell $i$~\cite{graham2019hover}. 
For ST images, we directly leverage the pre-computed segmentation annotations provided by the HEST-1k dataset, which are derived via CellViT~\cite{horst2024cellvit}~\cite{jaume2024hest}. 
These coordinates $\mathbf{c}_i$ serve as a critical input to our model for constructing spatial graphs.

\paragraph{\textbf{Gene selection.}}
Processing all genes is computationally expensive and may introduce noise. We employ a two-stage strategy to identify biologically informative genes.
First, to account for the inherent mean-variance relationship in transcriptomes (where highly expressed genes naturally have higher variance), raw counts are log-normalized. We then identify sample-specific \textit{High Variable Genes} (HVGs) by calculating a normalized dispersion score $z_g$. 
Instead of using raw variance, genes are grouped into bins based on their average expression. The normalized dispersion is defined as:
\begin{equation}
    z_g = \frac{d_g - \mu(B_g)}{\sigma(B_g)},
\end{equation}
where $d_g$ is the variance-to-mean ratio of gene $g$, and $\mu(B_g), \sigma(B_g)$ denote the mean and standard deviation of dispersion within a bin $B_g$. This step standardizes variance across different expression levels.

To construct a robust global gene set, we form a candidate pool $\mathcal{C}$ from the union of top-ranked HVGs across samples (excluding mitochondrial and ribosomal genes). 
To further filter out unstable or low-abundance noise, we compute the global mean expression $\mu_g$ and standard deviation $\sigma_g$ for each gene across the entire cohort. The final gene set $\mathcal{G}_{\text{final}}$ is derived by requiring genes to be both highly variable and highly expressed:
\begin{equation}
    \mathcal{G}_{\text{final}} = \text{Top}_K(\{\mu_g\}) \cap \text{Top}_K(\{\sigma_g\}).
\end{equation}
This intersection ensures the selected features capture significant biological heterogeneity (high $\sigma_g$) while ensuring reliable detection (high $\mu_g$) to avoid dropout artifacts.

\paragraph{\textbf{Gene expression normalization.}}
First, we perform the log1p normalization for Bulk and ST gene expression. The dynamic range of gene expression varies by orders of magnitude (e.g., thousands for marker genes vs. single digits for transcription factors). 
Without normalization, the regression loss function (e.g., MSE) is dominated by genes with large variances, causing the model to prioritize fitting highly expressed genes while ignoring subtle but critical patterns in low-expressed genes. This imbalance often results in an artificially inflated global Pearson correlation but poor gene-wise performance.

To address this, we apply gene-wise Z-score normalization to standardize the target distribution. Let $\mathrm{y}_{ij}$ be the log-transformed count of gene $j$ in patch $i$. We compute the global statistics $\mu_j$ (mean) and $\sigma_j$ (standard deviation) for each gene $j$ over the training set:
\begin{equation}
    \tilde{\mathrm{y}}_{ij} = \frac{\mathrm{y}_{ij} - \mu_j}{\sigma_j + \epsilon},
\end{equation}
where $\epsilon$ is a stability constant. This transformation rescales all genes to a standard normal distribution ($\mathcal{N}(0,1)$), ensuring they contribute equally to the gradient updates. During inference, predictions are denormalized using $\mu_j$ and $\sigma_j$ to recover the original biological scale.

\section{Implementation Details}
\label{appendix:implementation}

\subsection{Running Environment}

The experiments of bulk data training are conducted on Linux servers with 16 vCPU AMD EPYC 9K84 96-Core Processor, 150G RAM, and 2 H20-NVLink(96GB).
The experiments of data preprocessing and ST data training are conducted on Linux servers with AMD Ryzen Threadripper 3970X 32-Core Processor, 270G RAM, and 2 NVIDIA GeForce RTX 4090(24GB). Our method is implemented on PyTorch 2.6.0 and Python 3.10.16.

\subsection{Training Details}
For all models, we employed the Adam optimizer~\cite{kingma2014adam} and used a mixed loss function. To ensure training stability, the gradient norm was clipped to 1.0. The learning rate was tuned via a search over $\{1\text{e-}3, 5\text{e-}4, 1\text{e-}4, 1\text{e-}5\}$; while the optimal rate varied across modalities, $1\text{e-}4$ was adopted as the default. Note that different learning rates are impacted due to the complex optimization space of multiple instance learning. All models were trained for a maximum of 100 epochs. We observed that the bulk model typically required more epochs ($>30$) to reach convergence, whereas the ST model converged significantly faster ($>10$ epochs). Early stopping was implemented with a patience of 6 epochs. Additionally, the dropout rate was selected from $\{0, 0.1, 0.2\}$, and both the number of $k$-means clusters and $k$-nearest neighbors were set to 8. The weight $\lambda$ of cluster loss is searched from 0.1 to 0.5. 

For bulk data, the full training procedure for a single model requires approximately 20 hours in H20-NVLink Graphics cards. In contrast, for spatial transcriptomics data, completing all folds on large-scale datasets such as BRCA and PRAD takes roughly 12 hours on NVIDIA GeForce RTX 4090 cards. When performing cross-modal transfer learning, experiments can be conducted both with and without the utilization of Low-Rank Adaptation (LoRA).
\subsection{Reproduction details}

All experiments were conducted on hardware equipped with 24 Intel(R) Xeon(R) Gold 6348 CPUs @ 2.60GHz and A800 GPUs. For spatial transcriptomics data with different resolutions, we uniformly used the dataset provided by Hest-1k.

\textbf{Bulk transcriptomics prediction: }
We selected a subset of WSIs and their corresponding bulk sequencing data for each cancer type from the TCGA dataset, with the specific sample sizes of the training and test sets provided in Table \ref{tab:bulk_rna_sample_size}.
\begin{itemize}
    \item \textbf{HE2RNA:} During the training process, the maximum number of training epochs was set to 50, the patience value for the early stopping strategy was set to 20, and the learning rate was configured as $1 \times 10^{-7}$; all other hyperparameters were kept as the model’s default settings. In the feature extraction stage, ResNet50 was employed to extract features to compare with UNI extraction.
    \item \textbf{SEQUOIA:} The original model provided two options (UNI and ResNet50) for feature extraction, and we selected UNI for this step in our study. During model training, the learning rate was set to $1 \times 10^{-3}$, and the number of training epochs was 200, with all other parameters retained as their default values.
    \item \textbf{UNI+MLP:} We adopted the same feature extraction and tile aggregation schemes as SEQUOIA, while replacing the original ViT model in SEQUOIA with MLP. The core default architecture of this MLP regressor is as follows: the input layer is sequentially connected to a fully connected layer (Linear) with 1024 dimensions and a ReLU activation function, followed by another fully connected layer (Linear) with 1024 dimensions and a ReLU activation function, and finally an output layer with no activation function. We train this model with a learning rate $1 \times 10^{-3}$ and 100 epochs.
\end{itemize}

\begin{table*}[t]
  \caption{Ablation study on bulk transcriptomics prediction. We compare our base and transfer models under three configurations: standard (Trans+Cluster), without clustering (w/o Clu.), without Clustering and Trans (w/o Clu+Trans). O. and G. represent Overall and Gene Pearson correlations. The best result for each dataset is bolded.}
  \label{tab:full_ablation_final_bulk}
  \centering
  \small
  % 调整列间距，避免使用 extracolsep 的虚假对齐
  \setlength{\tabcolsep}{6pt} 
  \begin{tabular}{l cc cc cc cc}
    \toprule
    \textbf{Configuration} & \multicolumn{2}{c}{\textbf{BRCA}} & \multicolumn{2}{c}{\textbf{COAD}} & \multicolumn{2}{c}{\textbf{PRAD}} & \multicolumn{2}{c}{\textbf{LIHC}} \\
    \cmidrule(lr){2-3} \cmidrule(lr){4-5} \cmidrule(lr){6-7} \cmidrule(lr){8-9}
    & O. & G. & O. & G. & O. & G. & O. & G. \\
    \midrule
    BiTro.     & \textbf{0.743{\scriptsize$\pm$0.186}} & \textbf{0.091{\scriptsize$\pm$0.058}} & \textbf{0.946{\scriptsize$\pm$0.007}} & \textbf{0.017{\scriptsize$\pm$0.014}} & \textbf{0.831{\scriptsize$\pm$0.112}} & \textbf{0.043{\scriptsize$\pm$0.012}}& \textbf{0.844{\scriptsize$\pm$0.194}} & \textbf{0.117{\scriptsize$\pm$0.102}} \\
    w/o Trans.      & 0.711{\scriptsize$\pm$0.174} & 0.107{\scriptsize$\pm$0.065} & 0.937{\scriptsize$\pm$0.009} & 0.019{\scriptsize$\pm$0.008} & 0.821{\scriptsize$\pm$0.143} & 0.047{\scriptsize$\pm$0.024} & 0.823{\scriptsize$\pm$0.247} & 0.131{\scriptsize$\pm$0.074} \\
    w/o Clu+Trans.     & 0.708{\scriptsize$\pm$0.183} & 0.104{\scriptsize$\pm$0.062} & 0.933{\scriptsize$\pm$0.007} & 0.021{\scriptsize$\pm$0.009} & 0.814{\scriptsize$\pm$0.127} & 0.050{\scriptsize$\pm$0.022} & 0.811{\scriptsize$\pm$0.263} & 0.139{\scriptsize$\pm$0.093} \\
    \bottomrule
  \end{tabular}
\end{table*}
\begin{table*}[t]
  \caption{Ablation study on spatial transcriptomics prediction. We compare our base and transfer models under three configurations: standard (Norm+Cluster), without clustering (w/o Clu.), without normalization (w/o Norm.), and without clustering and normalization (w/o Norm+Clu.). O. and G. represent Overall and Gene Pearson correlations. The best result for each dataset is bolded.}
  \label{tab:full_ablation_final_st}
  \centering
  \small
  \setlength{\tabcolsep}{2.8pt} % 极致紧凑的列间距
  \begin{tabular*}{\textwidth}{@{\extracolsep{\fill}}ll cc cc cc cc cc}
    \toprule
    \textbf{Type} & \textbf{Config.} & \multicolumn{2}{c}{\textbf{BRCA}} & \multicolumn{2}{c}{\textbf{COAD}} & \multicolumn{2}{c}{\textbf{PRAD}} & \multicolumn{2}{c}{\textbf{HCC}} & \multicolumn{2}{c}{\textbf{CSCC}} \\
    \cmidrule(lr){3-4} \cmidrule(lr){5-6} \cmidrule(lr){7-8} \cmidrule(lr){9-10} \cmidrule(lr){11-12}
    & & O. & \hspace{-10pt}G. & O. & \hspace{-10pt}G. & O. & \hspace{-10pt}G. & O. & \hspace{-10pt}G. & O. & \hspace{-10pt}G. \\
    \midrule
    % Base Group
    \textbf{Base} & Norm+Clu. & 0.578{\tiny$\pm$0.11} & \hspace{-12pt}0.136{\tiny$\pm$0.13} & 0.566{\tiny$\pm$0.07} & \hspace{-12pt}0.094{\tiny$\pm$0.10} & \textbf{0.572{\tiny$\pm$0.15}} & \hspace{-12pt}0.100{\tiny$\pm$0.12} & 0.340{\tiny$\pm$0.01} & \hspace{-12pt}0.128{\tiny$\pm$0.03} & 0.574{\tiny$\pm$0.01} & \hspace{-12pt}0.231{\tiny$\pm$0.08} \\
    & w/o Clu.  & 0.587{\tiny$\pm$0.10} & \hspace{-12pt}0.154{\tiny$\pm$0.14} & 0.565{\tiny$\pm$0.07} & \hspace{-12pt}0.103{\tiny$\pm$0.09} & 0.502{\tiny$\pm$0.13} & \hspace{-12pt}0.186{\tiny$\pm$0.08} & 0.341{\tiny$\pm$0.02} & \hspace{-12pt}0.128{\tiny$\pm$0.03}& \textbf{0.667{\tiny$\pm$0.04}} & \hspace{-12pt}\textbf{0.429{\tiny$\pm$0.15}} \\
    & w/o Norm. & 0.618{\tiny$\pm$0.12} & \hspace{-12pt}0.120{\tiny$\pm$0.09} & 0.485{\tiny$\pm$0.13} & \hspace{-12pt}0.073{\tiny$\pm$0.07} & 0.526{\tiny$\pm$0.15} & \hspace{-12pt}0.091{\tiny$\pm$0.10} & 0.074{\tiny$\pm$0.03} & \hspace{-12pt}0.133{\tiny$\pm$0.04} & 0.202{\tiny$\pm$0.01} & \hspace{-12pt}0.250{\tiny$\pm$0.02} \\
    & w/o Norm+Clu. & 0.589{\tiny$\pm$0.09} & \hspace{-12pt}0.088{\tiny$\pm$0.09} & 0.456{\tiny$\pm$0.13} & \hspace{-12pt}0.104{\tiny$\pm$0.14} & 0.514{\tiny$\pm$0.12} & \hspace{-12pt}0.127{\tiny$\pm$0.07} & 0.081{\tiny$\pm$0.04} & \hspace{-12pt}0.134{\tiny$\pm$0.04} & 0.598{\tiny$\pm$0.06} & \hspace{-12pt}0.424{\tiny$\pm$0.16} \\
    \midrule
    % Transfer Group
    \textbf{Trans.} & Norm+Clu. & 0.588{\tiny$\pm$0.10} & \hspace{-12pt}0.162{\tiny$\pm$0.14} & 0.573{\tiny$\pm$0.06} & \hspace{-12pt}0.116{\tiny$\pm$0.10} & 0.516{\tiny$\pm$0.13} & \hspace{-12pt}0.204{\tiny$\pm$0.13} & 0.342{\tiny$\pm$0.01} & \hspace{-12pt}0.143{\tiny$\pm$0.01} & --- & \hspace{-12pt}--- \\
    & w/o Clu.  & \textbf{0.598{\tiny$\pm$0.10}} & \hspace{-12pt}\textbf{0.173{\tiny$\pm$0.14}} & \textbf{0.615{\tiny$\pm$0.06}} & \hspace{-12pt}\textbf{0.137{\tiny$\pm$0.10}} & 0.521{\tiny$\pm$0.13} & \hspace{-12pt}\textbf{0.214{\tiny$\pm$0.12}} & \textbf{0.342{\tiny$\pm$0.01}} & \hspace{-12pt}\textbf{0.144{\tiny$\pm$0.01}} & --- & \hspace{-12pt}--- \\
    & w/o Norm. & 0.586{\tiny$\pm$0.01} & \hspace{-12pt}0.155{\tiny$\pm$0.12} & 0.486{\tiny$\pm$0.13} & \hspace{-12pt}0.094{\tiny$\pm$0.11} & 0.519{\tiny$\pm$0.12} & \hspace{-12pt}0.198{\tiny$\pm$0.10} & 0.065{\tiny$\pm$0.04} & \hspace{-12pt}0.116{\tiny$\pm$0.06} & --- & \hspace{-12pt}--- \\
    & w/o Norm+Clu. & 0.586{\tiny$\pm$0.09} & \hspace{-12pt}0.155{\tiny$\pm$0.12} & 0.486{\tiny$\pm$0.13} & \hspace{-12pt}0.109{\tiny$\pm$0.11} & 0.489{\tiny$\pm$0.12} & \hspace{-12pt}0.198{\tiny$\pm$0.10} & 0.065{\tiny$\pm$0.04} & \hspace{-12pt}0.116{\tiny$\pm$0.06} & --- & \hspace{-12pt}--- \\
    \bottomrule
  \end{tabular*}
\end{table*}
\textbf{Spatial transcriptomics prediction: }
For spatial transcriptomics data at different resolutions, we modified the input interfaces of all models to be compatible with the unified dataset provided by Hest-1k. To enhance the robustness of the experiment, we adopted the leave-one-out cross-validation method for model training and evaluation in this study.
\begin{itemize}
    \item \textbf{HistoGene:} We set the maximum number of training epochs to 60. The learning rate ($\text{lr}$) was set to a default value of $5 \times 10^{-4}$. All model parameters were kept unchanged except for the maximum number of positional encodings ($n_{\text{pos}}$). While a value of 64 was sufficient for processing spatial transcriptomics datasets with small number of spots (e.g., Her2ST, CSCC), it failed to accommodate large datasets with abundant spot sequencing points (e.g., COAD). Thus, we increased the $n_{\text{pos}}$ parameter to 1024 to ensure full coverage of positional encoding requirements for large datasets.
    \item \textbf{Hist2ST:} We set the maximum number of training epochs to 300. The learning rate was set to a default value of $1\times 10^{-4}$. The code logic of this model is highly similar to that of HisToGene, and the same treatment of setting $n_{\text{pos}} = 1024$ was applied.
    \item \textbf{THItoGene:} We set the maximum number of training epochs ($\text{max\_epochs}$) to 150 epochs. The learning rate was set to a default value of $5 \times 10^{-4}$.
    The batch size was fixed at 1, and the same treatment of setting $n_{\text{pos}} = 1024$ was applied due to GPU memory constraints. Notably, when the loss curve of the model converges normally, the gene correlation on the test set sometimes exhibits an oscillating trend. The final result shows a high overall correlation but a low correlation, which may be due to the model primarily learning the overall gene distribution.
    
    \item \textbf{M2OST:} We set the maximum number of training epochs to 300. The learning rate was set to a default value of $1\times 10^{-4}$. And the batch size is 24.
\end{itemize}

\textbf{Cellular spatial transcriptomics prediction: }
For the subcellular resolution sequencing data generated via Xenium technology and its corresponding pseudo-ST data used in this experiment, we directly adopted the preprocessed dataset from Hest-1k. Following the spot aggregation protocol of Visium technology, pseudo-ST spots were aggregated into 224×224 pixel slices, with an actual physical size of $55\ \mu\text{m} \times 55\ \mu\text{m}$. Each data WSI was evenly split along the X-axis at a 4:1 ratio, and 5-fold cross-validation was performed for experimental evaluation, including the correlation between transcript counts from real cells and the model-predicted data.
\begin{itemize}
    \item \textbf{iStar:} This model was trained on pseudo-ST data generated by Xenium technology to produce super-pixel-level spatial transcriptomics data. For cellular aggregation, we selected only the cells present in the test set for aggregation, and the aggregation method entailed the direct summation of transcriptomic signals across all pixels within each cell. The core training hyperparameters of this experiment were configured as follows: the number of training epochs was set to 400, the learning rate was configured as 1e-4; additionally, five distinct initial states were set for the model during the initialization phase, which were used for subsequent model ensemble training.

    \item \textbf{GHIST:} We did not utilize the native HoVer-Net cell segmentation functionality provided by the original model; instead, we employed cell segmentation data generated through Xenium technology, reducing the substantial mismatches of cell segmentation and moving focus on gene prediction only. 
    Training and evaluation were conducted based on the inherent cell positions and cellular transcriptomics profiles of the dataset. The model adopted a learning rate of 1e-3 and a total of 50 training epochs, with all other parameters remaining unchanged.
\end{itemize}

\section{Ablation Study}
\label{appendix:ablation}

In this section, we provide a detailed breakdown of the ablation experiments to further elucidate the contribution of each module within the BiTro framework. We examine three specific configurations: the impact of transfer learning, the role of the clustering constraint, and the necessity of variance normalization. Results are shown in Table~\ref{tab:full_ablation_final_bulk} and Table~\ref{tab:full_ablation_final_st}.

We compared the performance of our full BiTro framework against a baseline model trained without transfer learning (denoted as "Base" for ST tasks and "w/o Trans." for bulk tasks). As shown in Table \ref{tab:bulk_final} and Table \ref{tab:st_comparison}, the transfer-enhanced models consistently outperform their non-transfer counterparts across nearly all cancer cohorts. For instance, in the bulk transcriptomics prediction task, pre-training on ST data allows the model to leverage learned spatial contexts, resulting in higher Pearson correlations and lower JS divergence. This validates our hypothesis that cross-modal knowledge transfer effectively bridges the gap between different data granularities.

Our experiments reveal that the utility of the K-means clustering module is sensitive to the granularity of the input supervision. In the context of bulk transcriptomics, removing the clustering module ("w/o Clu.") leads to a discernible performance drop. Since bulk data lacks high-precision labels, the phenotype clusters act as a proxy for tissue structure, offering a crucial fine-grained constraint that guides the Multiple Instance Learning (MIL) aggregator. In contrast, for Spatial Transcriptomics (ST) data, the "w/o Clu." configuration often achieves comparable or marginally better performance than the standard configuration. We attribute this to the fact that ST data already provides explicit spot-level supervision, rendering the implicit regularization from clustering redundant or, in some cases, overly distinctive for the learning objective.

Finally, we investigated the role of variance normalization in the regression head. The removal of this component ("w/o Norm.") results in a sharp decline in predictive accuracy, particularly in Gene Pearson correlation metrics. Without normalization, the loss function is disproportionately influenced by a small subset of genes with extremely high expression variance. Normalization ensures that the gradient descent process treats genes more equitably, allowing the model to capture expression trends across the entire transcriptome rather than overfitting to highly variable outliers.

\end{document}